\pdfoutput=1
\documentclass[10pt]{article}
\usepackage[preprint]{tmlr}
\usepackage{amsmath,amssymb,amsfonts}
\usepackage{graphicx,booktabs,caption,hyperref,url,float}
\usepackage{array}
\usepackage[T1]{fontenc}
\usepackage[capitalize,nameinlink]{cleveref}

\newcommand{\Dc}{D}                          
\newcommand{\Hr}{H_r}                         
\newcommand{\pmar}{(\Hr-\Dc)}                 
\newcommand{\lamz}{\lambda_0}                 
\newcommand{\dlamsq}{\lambda^2-\lamz^2}       
\newcommand{\Czero}{C_0}                      
\newcommand{\Cone}{C_1}                       

\title{Data Predictability Shapes Weibull Weight-Scale Growth in Transformer Training}

\author{\name Tiexin Ding\thanks{Independent Researcher. Email: \texttt{tiexinding@gmail.com}}}

\begin{document}
\maketitle

\begin{abstract}
A trained transformer's weight magnitudes can be summarized by a two-parameter Weibull distribution
whose shape $k\approx1.2$ is stable across layers and models, so the scale $\lambda$ carries most
training-induced movement. What corpus property sets how much $\lambda$ grows? Using the bigram
conditional entropy $\Dc = H(\text{next}\mid\text{prev})$, a training-free statistic computed before
training, we find across controlled corruption families a learning-rate-conditioned law, $\dlamsq =
\Czero(\eta) + \Cone(\eta)\,\pmar^{0.59}$, where $\Hr$ is a matched-budget shuffle baseline. The
convex exponent is inherited from an independently measured data-side saturation relation rather than
fitted directly to the growth curve. After removing the two per-$\eta$ coefficients, 23 runs spanning
an order of magnitude in learning rate collapse onto $\pmar^{0.59}$ with unit slope ($R^2 = 0.941$;
direct per-$\eta$ fits are weaker, $R^2 \approx 0.82$). Because $\Dc$ is computed before training, the
law is a forward predictor: an end-to-end self-validation recovers held-out within-family weight
growth with $5.7\%$ relative error. The readout holds at model and per-layer resolutions and across
two tested architectures, with the functional form preserved and only the coefficients changing. It
also marks its boundary: cross-corpus prediction over-predicts code, implicating redundancy as a
second axis of a broader $\Phi(\Dc, R, A, H)$ data-to-weight framework.
\end{abstract}

\section{Introduction}
The weights of a trained transformer are not an arbitrary bag of numbers: across a wide range of
models, the magnitudes of the weights in a linear layer follow a Weibull distribution, compactly
summarized by a shape parameter $k$ and a scale parameter $\lambda$ \citep{paper1_weibull}. The
shape is remarkably stable --- $k\approx1.2$ across layers and models --- so the single quantity that
moves during training is the scale $\lambda$. Recent work shows that $\lambda$ grows under AdamW as
the net of three forces --- injection, decay, and alignment --- and that this growth, not $\lambda$
itself, is where training dynamics live \citep{paper2_threeforce}. This raises a question that the
optimizer-side account leaves open: \textbf{what property of the training \emph{data} determines how
much $\lambda$ grows?} That same study makes the question concrete: it observed that the peak of
$\lambda(t)$ varies with training-data coherence and left a controlled data-side study to follow-up
work --- the present paper provides it.

We answer it with a single, training-free data statistic. Let $\Dc=H(\text{next}\mid\text{prev})$ be
the bigram conditional entropy of the corpus --- a measure of how predictable the next token is from
the previous one, computable from the raw data before any training. We find that the weight-scale
growth is an affine function of a fixed power of the \emph{predictability margin} $\pmar$,
\begin{equation}\label{eq:law}
  \dlamsq \;=\; \Czero(\eta) \;+\; \Cone(\eta)\,\pmar^{0.59},
\end{equation}
where $\lambda_0$ is the initial scale and $\Hr$ is the level the plug-in bigram entropy saturates
to under matched-budget shuffling (\cref{sec:limitations}). Two facts make this more than an
\emph{unconstrained} curve fit. First, the convex exponent is not selected by directly optimizing
the growth fit: it is tied to an independently measured data-side saturation exponent, $0.59=1/p$
with $p\approx1.69$ (\cref{sec:convex}), so the functional form is \emph{empirically constrained} by
a separate data-side relation rather than introduced as a free growth-curve exponent. Second, the relation
\textbf{collapses across learning rates}: after removing the two per-$\eta$ coefficients
$\Czero(\eta),\Cone(\eta)$, 23 runs spanning an order of magnitude in $\eta$ fall on the single curve
$y=\pmar^{0.59}$ with slope $1.00$ ($R^2=0.941$; \cref{sec:law}). Because those coefficients are
fitted per family, the collapse tests the \emph{shared exponent and unit slope} --- not the
per-$\eta$ predictive accuracy, which is weaker ($R^2\approx0.82$) and which we report explicitly in
\cref{sec:law}. The learning rate enters the law only through the two coefficients $\Czero(\eta)$ and $\Cone(\eta)$,
which follow power-law scalings over the four tested learning rates (\cref{eq:eta-scaling}).

Because the predictor's input $\Dc$ is computed from the corpus \emph{before training}, the law is a
genuine forward prediction: within the controlled corruption family, one can predict weight-scale
growth from a pre-training data statistic, and we verify this against actual training in a held-out,
end-to-end self-validation (\cref{sec:selfval}). This closes a loop between a
\textbf{data-side measurable} and a \textbf{weight-side dynamical quantity} --- a direction
complementary to existing data-evaluation methods, which mostly rely on downstream proxies or
heuristics. The ingredients themselves are classical --- conditional entropy is a statistical cousin of
perplexity, and the alignment quantity is a weight--gradient cosine. Our contribution lies in the
\emph{connection}: a standard data-side statistic and a Weibull weight-scale statistic are linked
through a controlled law that predicts a training outcome before training begins.

\textbf{A lens at the model and layer levels.} The same Weibull $(k,\lambda)$ ruler can be read at
two granularities. Over the whole model it gives a single scale-growth law tied to the corpus
statistic $\Dc$; per layer it yields blockwise $(k,\lambda)$ estimates that \emph{localize} where
predictable structure is written into the network. The law keeps the same functional form per block
with only block-specific coefficients (median per-block $R^2=0.84\approx$ the pooled $0.82$;
\cref{sec:robustness}). The shape $k$ stays nearly uniform across blocks (1.16--1.21) --- data
shapes the \emph{same kind} of distribution everywhere --- while the scale response is heterogeneous
and concentrated in the feed-forward blocks. Because the second moment
$\langle w^2\rangle=\lambda^2\,\Gamma(1+2/k)$ is additive under pooling and $k$ is uniform, the
pooled $\lambda^2$ is well approximated by the parameter-count-weighted average of the per-block
$\lambda^2$.

\textbf{Scope.} Our law is established \emph{within a corpus}, along a corruption gradient that
varies the next-token mapping while holding the corpus fixed; $\pmar$ is not a universal cross-corpus
currency, and we mark its boundary explicitly (code, whose low entropy is redundancy rather than
rich structure, departs from the law --- \cref{sec:limitations}). A second data dimension
(redundancy) and the cross-architecture and multi-epoch extensions are left as directions
(\cref{sec:limitations},~\cref{sec:discussion}).

\textbf{Contributions.}
\begin{enumerate}
  \item A \textbf{learning-rate-collapsing law} predicting Weibull weight-scale growth from a single
    training-free data statistic, with a convex exponent tied to an independently measured data-side
    relation rather than a free growth-curve parameter (\cref{sec:law}--\cref{sec:convex}).
  \item A \textbf{model-level and layer-level} account of weight-scale growth: the same
    $(k,\lambda)$ ruler exposes a nearly uniform distribution \emph{shape} ($k$) and a heterogeneous,
    feed-forward-dominated \emph{scale} ($\lambda$) response; the law holds at both granularities,
    and the pooled $\lambda^2$ form is well approximated by the parameter-weighted aggregate of the
    per-block laws (\cref{sec:robustness}).
  \item An \textbf{end-to-end self-validation}: predicting $\lambda$ from the data statistic alone
    matches measured $\lambda$ on held-out corruption levels (5.7\% relative error), with the
    residual localized to the training-side rung (\cref{sec:selfval}).
  \item An explicit \textbf{scope and future direction}: the law is established within controlled
    corruption families; cross-corpus, it predicts mapping-rich text but over-predicts code by a
    quantified margin, implicating redundancy as a second data dimension and motivating a broader
    multi-dimensional ($\Dc{+}R$) data framework (\cref{sec:limitations},~\cref{sec:discussion}).
\end{enumerate}

A summary of notation is provided in \cref{app:notation}.

\section{Background and Setup}\label{sec:background}

\textbf{Weibull weight scale.} Across transformers, the magnitudes of a linear layer's weights are well described by a Weibull distribution with shape $k$ and scale $\lambda$ \citep{paper1_weibull}. The shape is nearly universal ($k \approx 1.2$ for transmission-class matrices), so the moving quantity is the scale $\lambda$. We treat $\lambda$ as a readout of the weight distribution that can be applied at any granularity (\cref{sec:robustness}).

\textbf{Three-force dynamics.} Under AdamW, the squared scale evolves as the integral of a net force --- injection, decay, and alignment --- so that $\lambda^2(t) = \lamz^2 + \int (\text{net force})\,\mathrm{d}t$ \citep{paper2_threeforce}. The growth $\dlamsq$ is therefore an accumulated force balance: the injection term provides a data-independent floor, while the alignment term captures the coherent component of the update that depends on learnable data structure (a linearity we verify empirically in \cref{sec:convex}). This motivates modeling the growth $\dlamsq$ rather than $\lambda$ itself, and gives the law's intercept and slope (\cref{sec:law}--\cref{sec:convex}) an injection-floor / alignment-efficiency reading. Steady-state analyses characterize how the scale depends on optimizer hyper-parameters, $\lambda_{\text{steady}} \propto \sqrt{\eta/\lambda_{\text{wd}}}$ \citep{vanlaarhoven2017l2,fan2025robust,wang2024adamw}. The present study is complementary: it focuses on how training \emph{data} modulates the transient growth of $\lambda^2$ before saturation. That same study already flagged this dependence empirically --- the peak of $\lambda(t)$ shifts with training-data coherence --- and left a controlled data-side account to follow-up work, which this paper provides.

\textbf{The data-side variable.} We ask which pre-training property of the corpus sets the alignment term. This choice arose from a broad search over candidate data-side statistics, including higher-order and embedding-based notions of corpus structure; in this controlled setting, the bigram conditional entropy $\Dc = H(\text{next}\mid\text{prev})$ \citep{shannon1948,cover2006elements} emerged as the simplest statistic that is computable before training and predictive of weight-scale growth. Computed from adjacent token counts, $\Dc$ measures local next-token predictability: low $\Dc$ means the previous token strongly constrains the next, whereas high $\Dc$ means the next token remains uncertain even after conditioning on the previous one. The margin $\pmar$ then measures how much predictable local structure remains relative to a matched-budget shuffle baseline $\Hr$. The statistic is used here as a probe of local corpus structure, a statistical cousin of perplexity, rather than proposed as a new data-quality metric in itself (\cref{sec:limitations}).

\section{Method}\label{sec:method}

\textbf{Corruption gradient.} To isolate one data property --- the integrity of the local next-token mapping --- we hold the corpus fixed (wikitext) and vary only how much that mapping is destroyed. A corruption level $\rho \in [0,1]$ shuffles a fraction of within-sequence token positions; $\rho=0$ is the clean corpus and $\rho=1$ is fully scrambled. We summarize each corrupted corpus by two quantities computed from the data alone, before any training: its \textbf{bigram conditional entropy} $\Dc = H(\text{next}\mid\text{prev})$, estimated plug-in from the token counts of the first 2.4M tokens, and its \textbf{struct-retain fraction} $S$, the share of adjacent token pairs preserved relative to the clean corpus, which runs linearly from 1 to 0 and serves as the mechanistic structure variable in the derivation of \cref{sec:convex}. Because the plug-in bigram entropy saturates once most adjacent structure is destroyed, we use $\rho \le 0.6$ as the main non-saturated analysis regime and treat the saturated tail as a boundary case (\cref{sec:limitations}).

\textbf{Fixed anchors: $\lamz$ and $\Hr$.} $\lamz$ is the Weibull scale at the first checkpoint of each run --- measured, and identical across the corruption axis because all runs branch from a common checkpoint. Its value is a property of the model and initialization, not a universal constant: $\lamz = 0.0232$ for the continued-training runs and $0.0177$ for the from-scratch runs. Subtracting $\lamz^2$ isolates the growth and removes the initialization offset. $\Hr = 8.0$ is the value $\Dc$ saturates to under matched-budget shuffling --- a plug-in finite-sample ceiling of the bigram estimator at this token budget, not the information-theoretic marginal entropy (\cref{sec:limitations}). Neither $\lamz^2$ nor $\Hr$ is a free parameter: with the exponent set by the data-side saturation relation of \cref{sec:convex}, the growth fit estimates only the two coefficients $\Czero(\eta)$ and $\Cone(\eta)$.

\textbf{Estimating $\Dc$: bias and variance.} The margin $\pmar$ can be read as a matched-budget estimate of the predictive information of the bigram channel: both terms use the same plug-in estimator and token budget, so much of the finite-sample bias is shared --- though not perfectly, since the occupied bigram support changes with corruption. Its sampling \emph{variance} is small: across twelve disjoint 2.4M-token subsamples the standard error of $\Dc$ stays below $0.011$ bits (under 0.2\%) at every corruption level --- smaller than the plotting markers --- and a moving-block bootstrap agrees, so the law's residual scatter is not an artifact of $\Dc$-estimation noise.

\textbf{Weibull weight-scale readout.} We fit a Weibull distribution to the magnitudes of the transmission-class weight matrices --- the attention output projection and the two feed-forward matrices --- and report its scale $\lambda$. Unless stated otherwise, $\lambda$ is a single \textbf{model-level} fit: the magnitudes of all such matrices across all layers are pooled into one distribution and fit once, by a log-log Weibull-plot regression over the 10--90\% quantile band. We exclude the query/key/value projections, whose magnitudes are not Weibull (a distinct, selection-type regime \citep{paper1_weibull}), and embeddings, norms, and biases. The same $(k,\lambda)$ lens applies per layer (\cref{sec:robustness}); the pooled $\lambda^2$ is well approximated by the parameter-count-weighted aggregate of the per-block $\lambda^2$ values, because second moments add and $k$ varies only weakly across blocks. Our pooled $k \approx 1.17$ is consistent with the per-block median $k \approx 1.20$ obtained by fitting each block separately; we use the pooled readout as the model-level summary that pairs with the corpus-level statistic $\Dc$.

\textbf{Training protocol.} All runs continue-train Pythia-70m with weight decay $\lambda_{\text{wd}}=0.01$, a linear warm-up over the first 200 steps, and cosine decay thereafter; the main analysis uses $\eta=3\text{e-}4$, with four learning rates spanning $1\text{e-}4$ to $1\text{e-}3$ and three seeds at the reference $\eta$. Both data- and weight-side quantities are used as established probes: $\Dc$ as a pre-training data statistic (computed from token counts \emph{before} training, not from model outputs), and the alignment term as a weight--gradient coherence measure (\cref{sec:limitations}).

\textbf{Interpretation and boundary of $\Dc$.} The corruption suite fixes the reading of $\Dc$ as local mapping predictability: shuffling positions and replacing tokens both destroy the next-token mapping and raise $\Dc$, while repetition --- which tiles real text and preserves each local mapping --- instead lowers $\Dc$ through redundancy. Low $\Dc$ can therefore arise from two different sources: meaningful local mappings in natural text, or cheap predictability from repetition and templating. The law below targets the first regime; the second becomes the boundary that motivates the redundancy dimension in \cref{sec:limitations}.

\section{A Conditional-Entropy Law for Weight-Scale Growth}\label{sec:law}

\subsection{Setup and the per-learning-rate relationship}

We continue-train Pythia-70m on a family of corpora obtained by progressively corrupting a single base corpus (\cref{sec:method}). The only quantity varied along the family is the integrity of the local next-token mapping, summarized by the bigram conditional entropy $\Dc = H(\text{next}\mid\text{prev})$. For each run we record the Weibull scale parameter $\lambda$ along the trajectory and take its initial value $\lamz$ at the first checkpoint. Because all runs in a family branch from a common checkpoint, $\lamz$ is identical across the corruption axis ($\lamz = 0.0232$ for every corruption level; \cref{sec:method}). The growth $\dlamsq$ therefore isolates what training \emph{added} and removes the initialization offset, which is essential for comparing runs that start from different checkpoints (\cref{sec:robustness}).

Holding the learning rate $\eta$ fixed and sweeping the corruption level, the weight-scale growth is an affine function of a single data-side variable --- the predictability margin $\pmar$ raised to a fixed exponent:
\begin{equation}
  \dlamsq \;=\; \Czero(\eta) \;+\; \Cone(\eta)\,\pmar^{0.59}.
  \label{eq:law-affine}
\end{equation}
Here $\Hr \approx 8.0$ is the empirical saturation value that $\Dc$ approaches when the local mapping is destroyed --- the matched-budget shuffle baseline (\cref{sec:method} and \cref{sec:limitations}). The exponent $0.59 = 1/p$ is tied to the independently measured data-side saturation relation of \cref{sec:convex}. The intercept $\Czero(\eta)$ is an \emph{injection floor} --- the residual growth when the predictive margin approaches zero ($\pmar \to 0$) --- and the slope $\Cone(\eta)$ is the \emph{alignment efficiency} with which predictable local structure is converted into weight-scale growth. With $\Hr$ measured and the exponent set from the data-side relation, \cref{eq:law-affine} fits only two per-$\eta$ coefficients, $\Czero(\eta)$ and $\Cone(\eta)$.

\begin{figure}[t]\centering
  \includegraphics[width=0.78\linewidth]{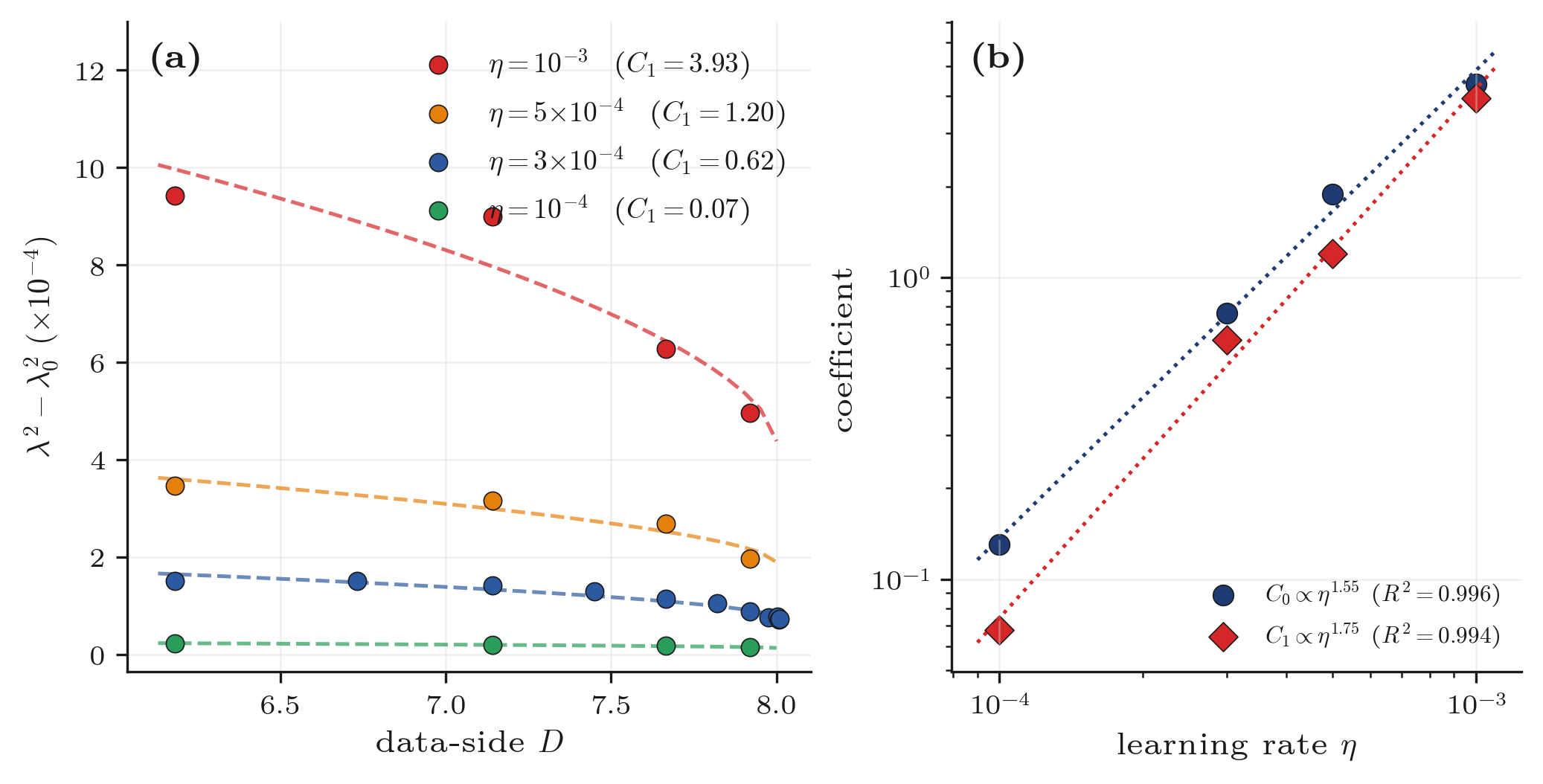}
  \caption{\textbf{Learning-rate scaling of the coefficients.} (a) per-$\eta$ gradients of
    \cref{eq:law-affine}; (b) the fitted coefficients follow power laws $\Czero\propto\eta^{1.55}$
    ($R^2=0.996$) and $\Cone\propto\eta^{1.75}$ ($R^2=0.994$). Four-point grid (two single-seed): a
    scaling description, not an established law.}
  \label{fig:eta}
\end{figure}

\Cref{eq:law-affine} holds across four learning rates spanning an order of magnitude. The fitted coefficients (\cref{tab:eta-coeffs}) vary systematically with $\eta$, following power laws
\begin{equation}
  \Czero \propto \eta^{1.55}\;(R^2=0.996), \qquad
  \Cone \propto \eta^{1.75}\;(R^2=0.994),
  \label{eq:eta-scaling}
\end{equation}
shown in \cref{fig:eta}. These scalings are measured over a four-point sweep (two of the four $\eta$ are single-seed) and should be read as empirical $\eta$-conditioning rather than a precise asymptotic law. The connection to optimizer dynamics is interpretive: the three-force account models $\lambda^2$ growth as an accumulated balance of injection, alignment, and decay, and the two coefficients are the data-experiment projections of that balance --- $\Czero$ tracking the data-independent injection floor and $\Cone$ the alignment channel. Because $\eta$ rescales every per-step update, it shifts both accumulated contributions; the resulting non-trivial exponents --- lying between the linear and quadratic scalings that single-term intuitions would suggest --- are consistent with such a multi-term balance, but do not by themselves constitute a quantitative test of it.

\begin{table}[t]
\centering
\caption{Fitted coefficients of \cref{eq:law-affine} per learning rate (convex form, exponent 0.59 fixed). $n$ = number of seeds; $\eta=3\text{e-}4$ is the seed-replicated reference (\cref{sec:robustness}). $\Czero$ and $\Cone$ are in units of $10^{-4}$.}
\label{tab:eta-coeffs}
\begin{tabular}{cccc}
\toprule
$\eta$ & $\Czero$ & $\Cone$ & $n$ \\
\midrule
1e-3 & 4.38 & 3.93 & 1 \\
5e-4 & 1.89 & 1.20 & 2 \\
3e-4 & 0.76 & 0.62 & 3 \\
1e-4 & 0.13 & 0.07 & 1 \\
\bottomrule
\end{tabular}
\end{table}

\subsection{\texorpdfstring{Collapse to an $\eta$-normalized data law}{Collapse to an eta-normalized data law}}

\Cref{eq:law-affine} implies that the learning rate enters only through the two coefficients $\Czero(\eta)$ and $\Cone(\eta)$. We therefore test the stronger normalized prediction
\[
  \bigl(\dlamsq - \Czero(\eta)\bigr)/\Cone(\eta) = \pmar^{0.59},
\]
which should hold for all runs if the functional form is shared across learning rates.

\begin{figure}[t]\centering
  \includegraphics[width=0.82\linewidth]{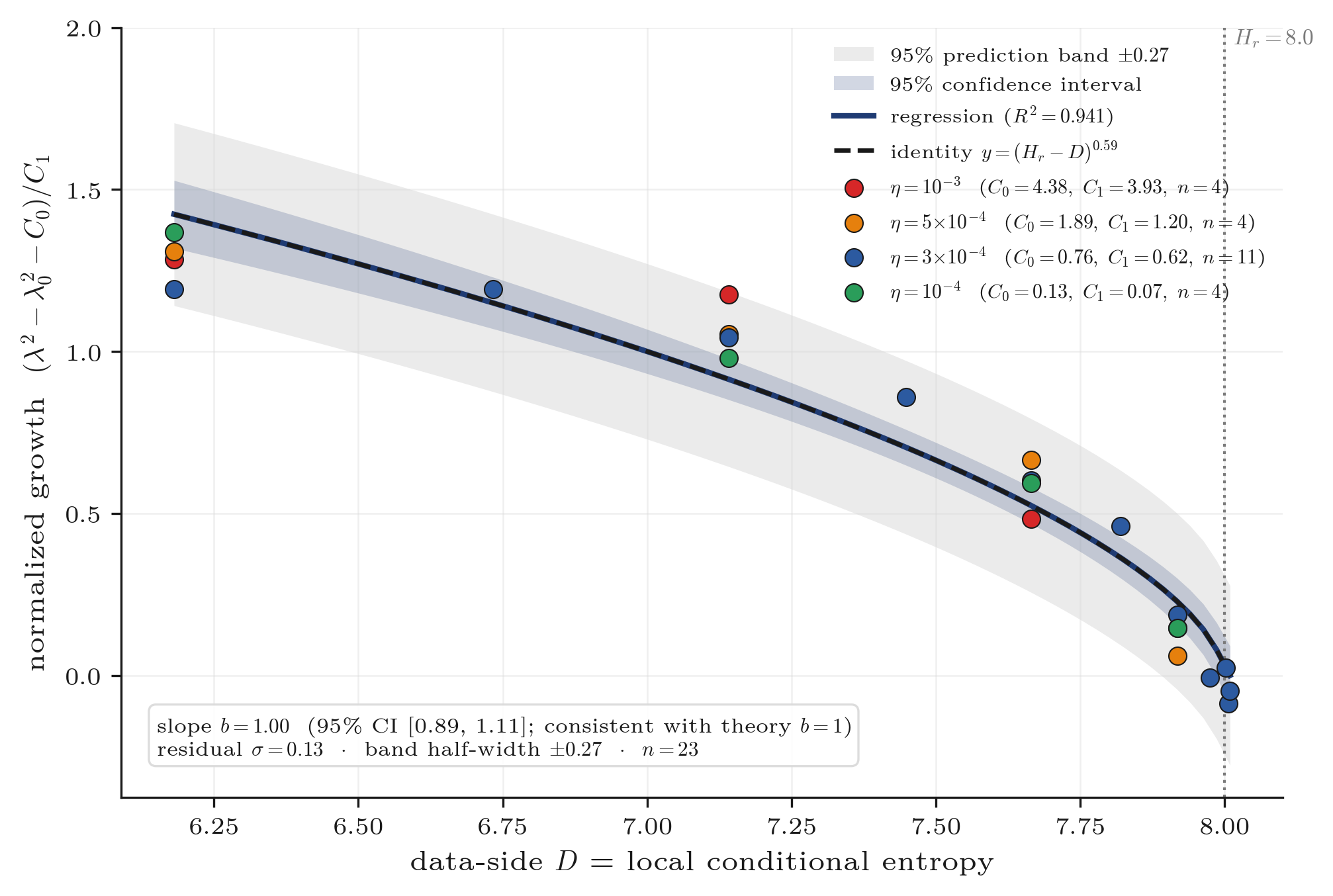}
  \caption{\textbf{Learning-rate collapse.} After removing the two per-$\eta$
    coefficients $\Czero(\eta),\Cone(\eta)$, the 23 $(\eta,\text{corruption})$ runs across four
    learning rates ($\eta=10^{-4}$ to $10^{-3}$) fall on the parameter-free identity
    $y=\pmar^{0.59}$ ($R^2=0.941$; regression slope $1.00$, 95\% CI $[0.89,1.11]$; residual
    $\sigma=0.13$, 95\% prediction band $\pm0.27$, $n=23$). The four $\eta$ families interleave
    rather than separating by colour --- the signature that the functional form is shared after
    normalization. Per-$\eta$ predictive accuracy is weaker ($R^2\approx0.82$; \cref{sec:law}).}
  \label{fig:collapse}
\end{figure}

\Cref{fig:collapse} shows this collapse for 23 runs spanning four tested learning rates. After normalization, the points fall close to the parameter-free identity $y = \pmar^{0.59}$, with $R^2 = 0.941$. A regression on the collapsed points gives slope $1.00$ with 95\% confidence interval $[0.89, 1.11]$, residual scatter $\sigma = 0.13$, and a 95\% prediction band of $\pm 0.27$. The learning-rate families interleave rather than forming separate bands, indicating that $\eta$ primarily rescales the same data law through $\Czero(\eta)$ and $\Cone(\eta)$.

This collapse should be read as a test of the shared functional form, not as evidence of exact per-$\eta$ prediction. Because $\Czero(\eta)$ and $\Cone(\eta)$ are fitted within each learning-rate family, the key evidence is the unit slope, small residual scatter, and interleaving across $\eta$, rather than the $R^2$ value alone. Within a single $\eta$, \cref{eq:law-affine} is less exact ($R^2 \approx 0.82$); \cref{sec:robustness} shows that this residual is not primarily seed noise or $\Dc$-estimation noise, but reflects the structure a one-dimensional bigram proxy leaves out --- higher-order organization, redundancy, and architecture-dependent responses.

\subsection{The law holds throughout the stable phase of training, not only at the endpoint}

\begin{figure}[t]\centering
  \includegraphics[width=0.82\linewidth]{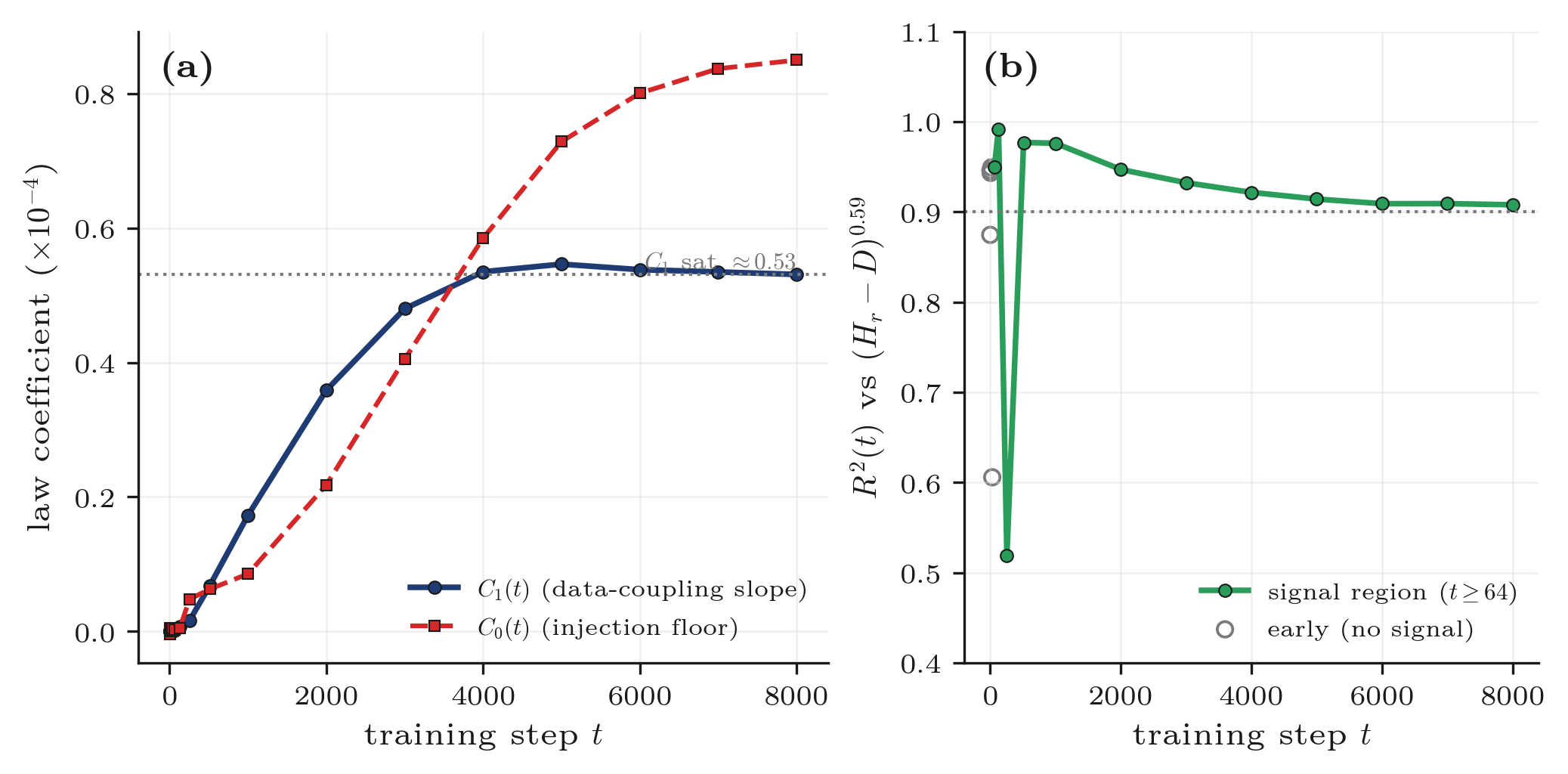}
  \caption{\textbf{The law holds throughout the stable phase, not only at the endpoint.} Refitting
    \cref{eq:law-affine} at each step (linear step axis). (a) the data-coupling slope $\Cone(t)$ rises
    from near zero and saturates (annotated $C_1$ sat), while the injection floor $\Czero(t)$
    accumulates; (b) the per-step fit $R^2(t)$ stays high ($0.91$--$0.99$) throughout the stable phase.
    The single dip to $R^2\approx0.52$ near step 250 is the warm-up snapshot, where $\eta$ ramps
    sharply while data coupling has barely begun --- a schedule transient, not a breakdown of the law.
    Filled: signal region $t\geq64$; open: early no-signal steps.}
  \label{fig:law-step}
\end{figure}

\Cref{eq:law-affine} relates the \emph{total} growth over a run to the data, with no explicit step index. To check that this is a property of the trajectory rather than an endpoint artifact, we refit \cref{eq:law-affine} at each recorded step $t$, giving step-resolved coefficients $\Czero(t), \Cone(t)$ (\cref{fig:law-step}). Through the stable phase --- after the learning-rate warm-up, once the scale has begun to move --- the relation holds throughout, with a per-step $R^2 = 0.91\text{--}0.99$ all the way to the endpoint. The coupling is carried by the coefficients: the data-coupling slope $\Cone(t)$ rises from near zero and saturates, while the injection floor $\Czero(t)$ accumulates monotonically --- the accumulated alignment work and injection of the three-force account, not the instantaneous learning rate or the fixed $\lamz^2$. This is the transient rise phase, not a steady state --- the relaxation timescale $\tau \approx 1/(\eta\lambda_{\text{wd}})$ far exceeds the 8000-step run \citep{wang2024adamw} --- so the law's cleanness here reflects the smoothness of the post-warm-up schedule, not equilibrium.

The one departure is a single snapshot at the end of warm-up (around step 250), where the per-step fit drops to $R^2 \approx 0.52$. Warm-up is the only place where the learning rate changes abruptly: $\eta$ ramps from near zero to its peak over $\sim\!200$ steps --- two orders of magnitude, far faster than anywhere else in the run --- while the data-coupling slope $\Cone(t)$ has barely turned on. At that instant the weight scale is driven by the steeply rising schedule rather than by data structure, so a single fit of $\dlamsq$ against $\pmar^{0.59}$ is momentarily poor. Once $\eta$ passes its peak and varies slowly, the fit recovers and stays high throughout the rest of training. The dip is therefore a schedule-induced transient, not a breakdown of the law.

\section{Why the Form Is Convex: A Mechanistic Account}\label{sec:convex}

\begin{figure}[t]\centering
  \includegraphics[width=0.95\linewidth]{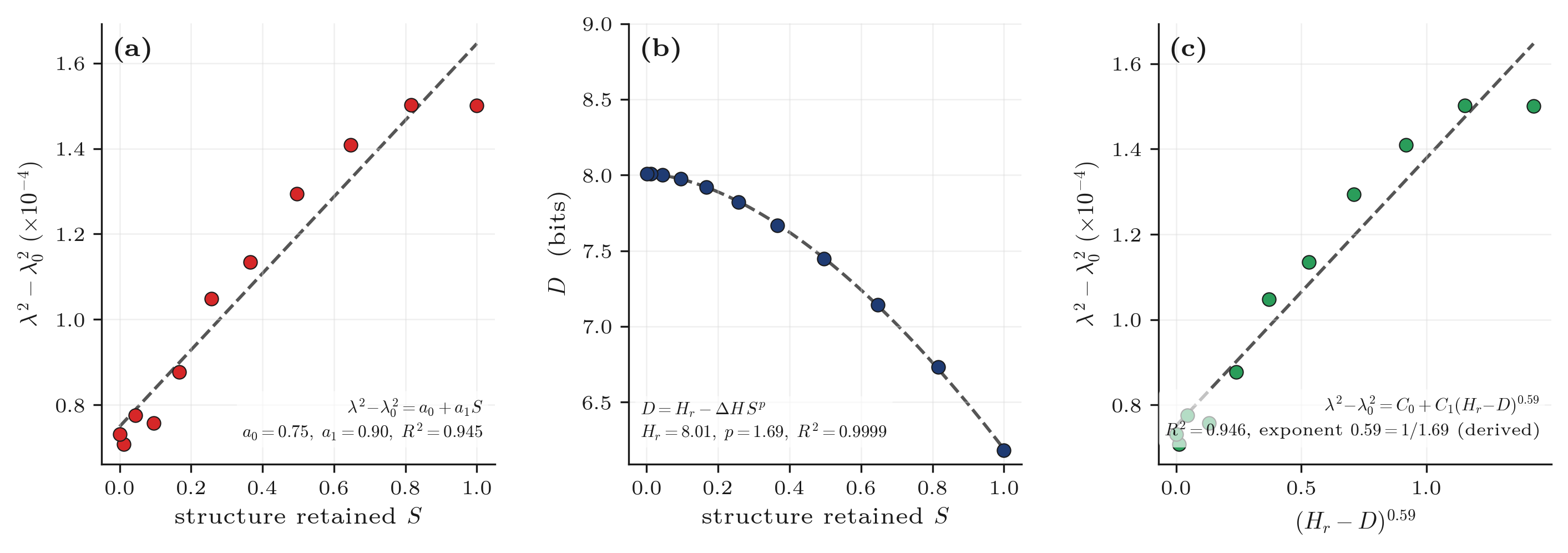}
  \caption{\textbf{Why the form is convex (derivation, \cref{sec:convex}).} (a) training side: growth
    is linear in retained structure $S$ ($R^2=0.945$); (b) data side: bigram entropy saturates,
    $\Dc=\Hr-\Delta H\,S^{1.69}$ ($R^2=0.9999$, $p=1.69$); (c) composing the two and eliminating $S$
    gives the convex law in $\pmar^{0.59}$ ($R^2=0.946$), with exponent $0.59=1/p$ inherited from the
    data-side saturation fit. Single corpus and corruption family.}
  \label{fig:deriv}
\end{figure}

The convex form and reciprocal exponent of \cref{eq:law} are derived, not fitted to the growth data. The derivation composes two independently measured relationships --- one each on the training and data sides --- through a single experiment-internal bridge variable, $S \in [0,1]$, the \emph{struct-retain fraction} (the fraction of adjacent token pairs preserved relative to the clean corpus, decreasing from $1$ clean to $0$ scrambled). $S$ is a readout of the corruption knob, \textbf{not a property of data} (a natural corpus has no $S$); it cancels from the final law, which is stated entirely in the training-free $\Dc$. Because $S$ and $\Dc$ both follow from the same corruption level $\rho$, the data-side relation below is a designed co-variation, not an independent discovery. We walk the construction through \cref{fig:deriv} and leave the full $\rho$--$S$--$\Dc$ tables, the determination of $\Hr$, and the exponent's robustness to \cref{app:construction}. (Here $a_0,a_1$ are the training-side fit coefficients in $S$; their relation to the law's $\Czero,\Cone$ is established by the composition below, \cref{eq:compose}.)

\emph{Training side} (\cref{fig:deriv}a). Growth is linear in retained structure,
\begin{equation}\label{eq:train-linear}
  \dlamsq = a_0 + a_1\,S, \qquad R^2 = 0.945,
\end{equation}
with a non-zero intercept: $a_0$ is the data-independent injection floor of the three-force decomposition \citep{paper2_threeforce} and $a_1 S$ the alignment contribution (coefficient values and the through-origin test are in \cref{app:construction}).

\emph{Data side} (\cref{fig:deriv}b), \textbf{the source of the convexity}. The training-free bigram entropy saturates in $S$,
\begin{equation}\label{eq:data-saturate}
  \Dc = \Hr - \Delta H \, S^{p}, \qquad p \approx 1.69 \ \ (R^2 = 0.9999),
\end{equation}
with ceiling $\Hr \approx 8.0$ and range $\Delta H \approx 1.82$. The exponent $p$ is an empirical fit, well-constrained and calibrated on this corruption family rather than claimed universal (\cref{app:construction}). Because $p > 1$, $\Dc$ loses sensitivity as structure is destroyed ($\mathrm{d}\Dc/\mathrm{d}S \to 0$ as $S \to 0$): the bigram ruler compresses at its degenerate, scrambled end.

\emph{Composition} (\cref{fig:deriv}c). Inverting \cref{eq:data-saturate} for $S$ and substituting into \cref{eq:train-linear} forces a power law in the predictability margin with the \emph{reciprocal} exponent,
\begin{equation}\label{eq:compose}
  \dlamsq = \Czero + \Cone\,\pmar^{1/p}, \qquad 1/p = 0.59,
\end{equation}
where $S$ has cancelled and the coefficients map onto the training-side fit ($\Czero = a_0$, $\Cone = a_1/\Delta H^{0.59}$, worked out in \cref{app:construction}). The conditional is a genuine result: \emph{given} a linear training side and a power-law data side, the composed exponent is fixed at $1/p$, not tuned to the growth data --- but both premises are themselves empirical fits, so the form is derived \emph{conditional on} two measured relationships, not from first principles.

The convexity is therefore a property of the data-side ruler, not of the training dynamics: training is linear in retained structure, and the bend enters only through the saturation of static bigram entropy. Read in reverse, this is why $\lambda$ keeps resolving structure at the degenerate end where $\Dc$ has already flattened --- the weight-scale readout retains resolution where the static bigram ruler loses it.

\section{End-to-End Self-Validation}\label{sec:selfval}

\begin{figure}[t]\centering
  \includegraphics[width=0.85\linewidth]{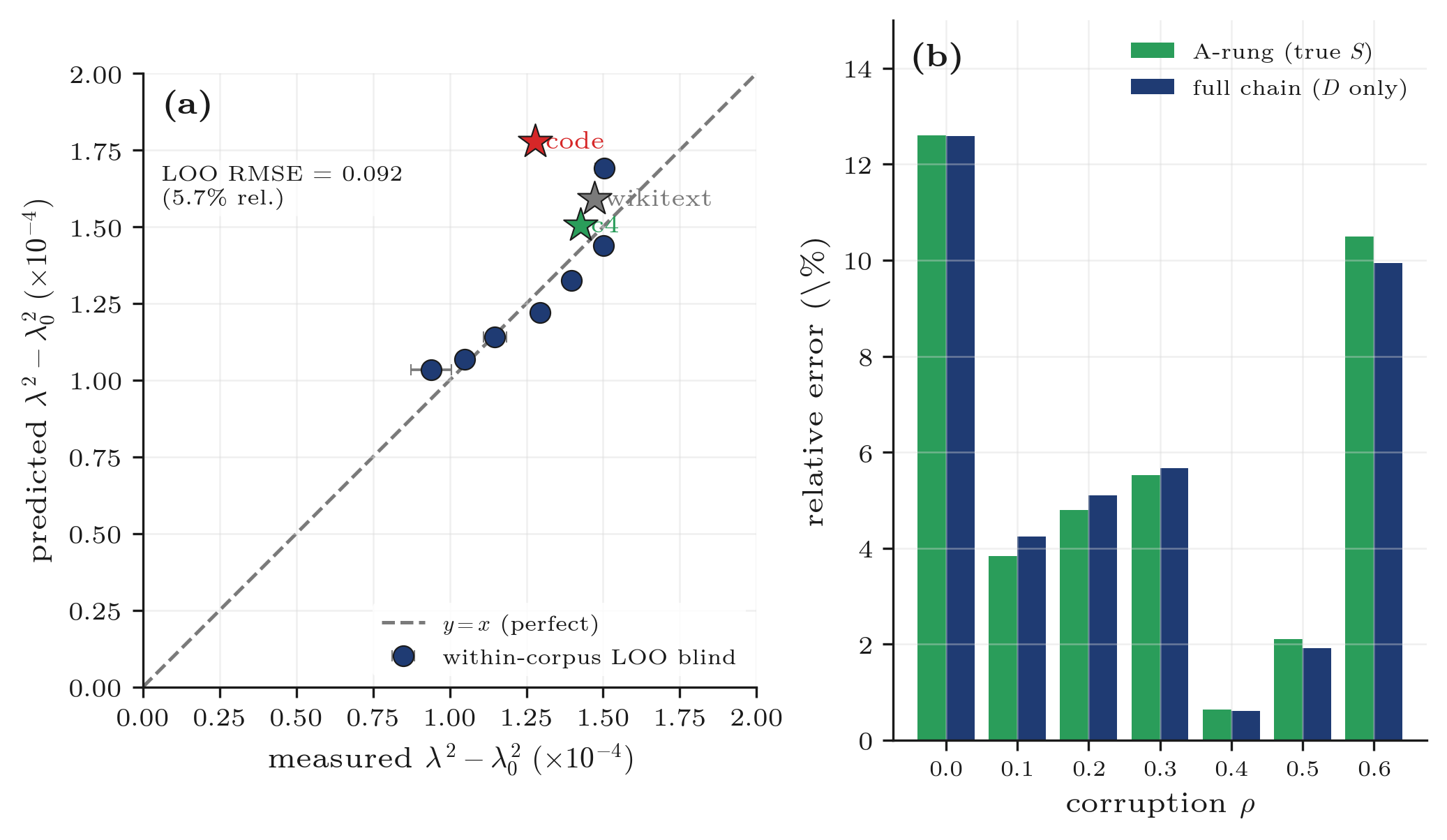}
  \caption{\textbf{End-to-end self-validation (\cref{app:selfval}).} The prediction chain is
    $\Dc\to S\to\lambda^2$: from the training-free statistic $\Dc$ we recover the struct-retain
    fraction $S$ via the data-side saturation relation, then apply the training-side fit. \emph{LOO}
    (leave-one-out): each corruption level is predicted from a fit on the other six, so the predicted
    point never enters its own fit. \textbf{(a)} predicted vs.\ measured growth: within-corpus LOO
    blind prediction (circles; LOO RMSE $0.092$, $5.7\%$) tracks the identity $y=x$; the held-out
    c4 ($\text{obs}-\text{pred}=-0.08$) and the in-family wikitext reference ($-0.12$) are predicted well,
    but code is over-predicted by $0.50$ ($\text{obs}-\text{pred}=-0.499$) --- the cross-corpus
    failure the scope predicts. \textbf{(b)} per-level relative error, decomposed: \emph{A-rung (true
    $S$)} uses the measured $S$ directly (isolating the training-side fit), while \emph{full chain
    ($\Dc$ only)} recovers $S$ from $\Dc$ first. The two are nearly equal at every $\rho$, so the
    $\Dc\!\to\!S$ inversion adds essentially nothing and the residual is the linear-in-$S$
    training-side error --- largest at the $\rho=0$ and $\rho=0.6$ ends, where the convex curvature is
    strongest.}
  \label{fig:selfval}
\end{figure}

Because $\Dc$ is computed from the raw corpus \emph{before} any training, the law can be tested as a genuine forward prediction: compute $\Dc$ from a corpus, predict $\dlamsq$ from the data alone, then train and measure $\dlamsq$ from the Weibull fit of the weights --- the same quantity obtained two independent ways. The prediction chain is $\Dc\to S\to\lambda^2$: invert the data-side saturation relation to recover the struct-retain fraction $S$ from the measured $\Dc$, then apply the training-side fit (\cref{fig:selfval}). The strength of the test rests on one condition: the predicted data does not enter the coefficient fit. The full protocol and tables are in \cref{app:selfval}.

\textbf{Within-corpus (held-out corruption levels).} Leave-one-out across the corruption family --- each level predicted from a fit on the other six, so a level never enters its own fit --- gives a blind RMSE of $0.092\times10^{-4}$, a $5.7\%$ relative error, against an in-sample RMSE of $0.054$ (\cref{fig:selfval}a). The full chain (from $\Dc$) and the training-side rung (from the measured $S$) agree to three decimals (\cref{fig:selfval}b), so the $\Dc\!\to\!S$ inversion contributes essentially no error; the residual is the linear-in-$S$ training-side scatter, largest at the corruption-axis ends where the convex curvature is strongest.

\textbf{Cross-corpus (held-out corpora).} With coefficients fit only on the wiki corruption family, the chain predicts mapping-rich held-out text well --- c4 within $-0.08$ --- but over-predicts code by $0.50$ in magnitude ($\text{obs}-\text{pred}=-0.499$, $\approx 6\times$ the within-corpus error). Code's low $\Dc=5.30$ is read as ``highly predictable,'' yet its low entropy comes from redundancy rather than a rich mapping, so it grows far less. This makes the \cref{sec:limitations} boundary quantitative: the predictor succeeds where the mapping interpretation of $\Dc$ holds and fails by a wide, well-defined margin where redundancy takes over --- the cross-corpus failure the scope predicts.

\section{Robustness across Seed, Architecture, and Layer}\label{sec:robustness}

\subsection{Seeds}
At the reference learning rate, we repeated four corruption levels, $\rho \in \{0,0.2,0.4,0.6\}$, over three seeds. The resulting variation in $\dlamsq$ is small: the coefficient of variation is 3.9\% across replicated settings (\cref{fig:crossseed}). Averaging over seeds changes the per-$\eta$ convex-law fit only modestly, from $R^2 = 0.909$ to $R^2 = 0.936$.

Thus, seed variation contributes to the scatter but does not dominate it. The remaining residual is better interpreted as structure left outside a one-dimensional bigram proxy: $\Dc$ captures local predictability, while higher-order organization, redundancy, and architecture-dependent responses can still perturb the growth. This reading is consistent with the collapse in \cref{sec:law}, which shows a shared functional form across learning rates even though each single-$\eta$ family retains residual scatter.

\subsection{Architecture}
To test whether the data law is tied to a particular architecture, we repeat the corruption experiment with two 70M-scale models trained \emph{from scratch} under the same data and optimization budget. We train Pythia-70m and a Llama-style 70m variant built for this study, since no official Llama model exists at this scale. Both are trained on the same seven corruption levels $\rho \in \{0,0.1,\dots,0.6\}$ for 8000 steps ($\approx$98M tokens, batch 24, sequence length 512), with weight decay $\lambda_{\text{wd}}=0.01$ and the reference learning rate $\eta=3\text{e-}4$ under the same warm-up/cosine schedule. The only difference is the feed-forward and normalization blocks: Pythia uses GELU feed-forward layers with LayerNorm, the Llama-style model SwiGLU gated feed-forward layers with RMSNorm.

\begin{figure}[t]\centering
  \includegraphics[width=0.82\linewidth]{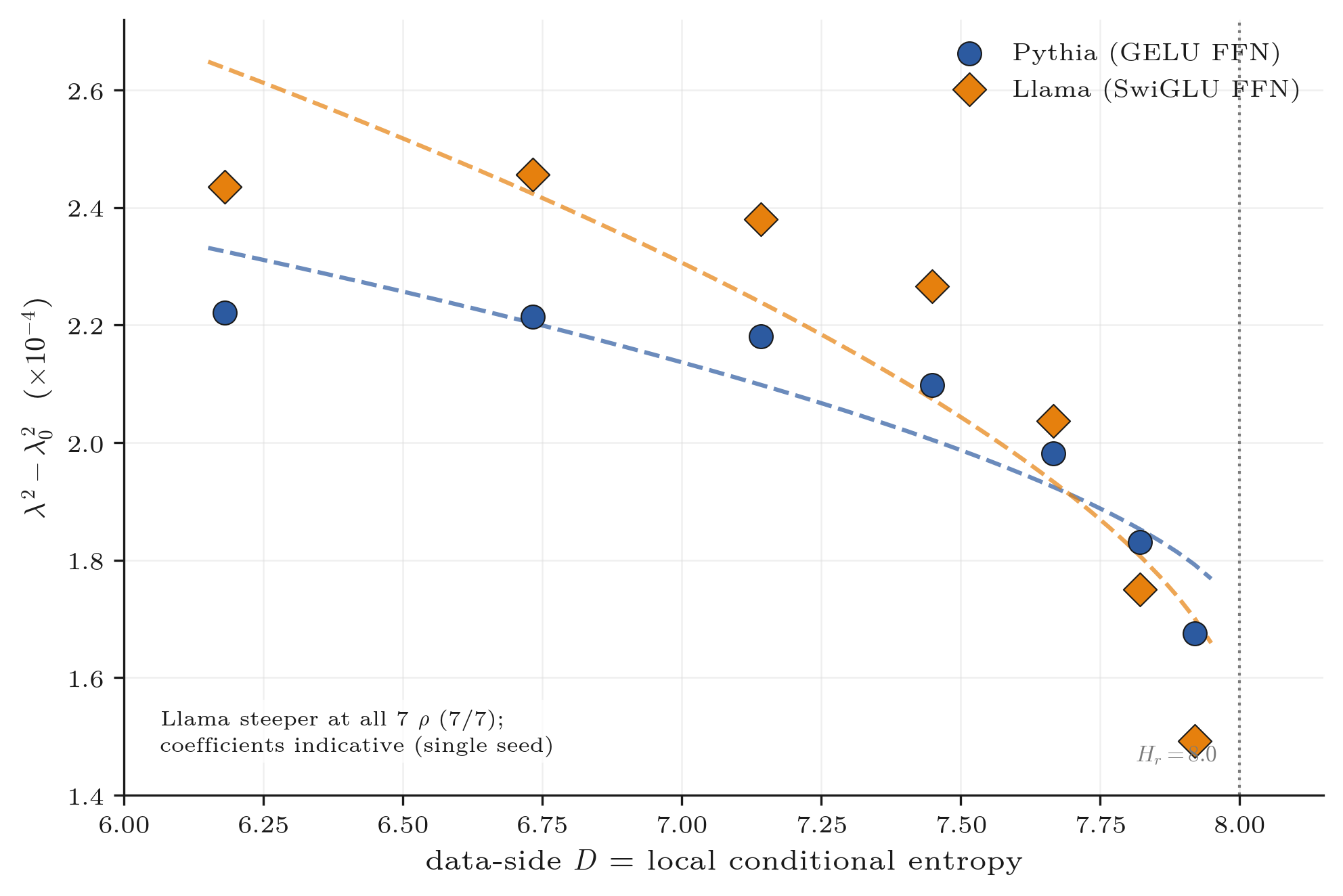}
  \caption{\textbf{Same convex form across architectures.} Pythia (GELU feed-forward) vs Llama
    (SwiGLU feed-forward), trained from scratch on the same seven-level corruption family: both
    follow \cref{eq:law} with the same sign and convex shape ($R^2=0.83/0.81$); the Llama-style model
    has the steeper trend across $\rho$ (slope $\Cone=0.78$ vs $0.45$). Coefficients are indicative
    (single seed each); the magnitude difference is not precisely estimated (\cref{sec:limitations}).}
  \label{fig:arch-compare}
\end{figure}

Both architectures follow the same law. Weight-scale growth $\dlamsq$ decreases monotonically as the local next-token mapping is degraded, and \cref{eq:law} fits the seven points with $R^2 = 0.83$ (Pythia) and $0.81$ (Llama) (\cref{fig:arch-compare}). The difference is not in the sign or convex form but in the coefficients: the Llama-style model's growth falls more steeply across the corruption axis, with fitted slope $\Cone = 0.78$ vs $0.45$ for Pythia. Each architecture is a single seed, so we report this coefficient difference as indicative rather than precisely estimated (\cref{sec:limitations}).

\begin{figure}[t]\centering
  \includegraphics[width=0.95\linewidth]{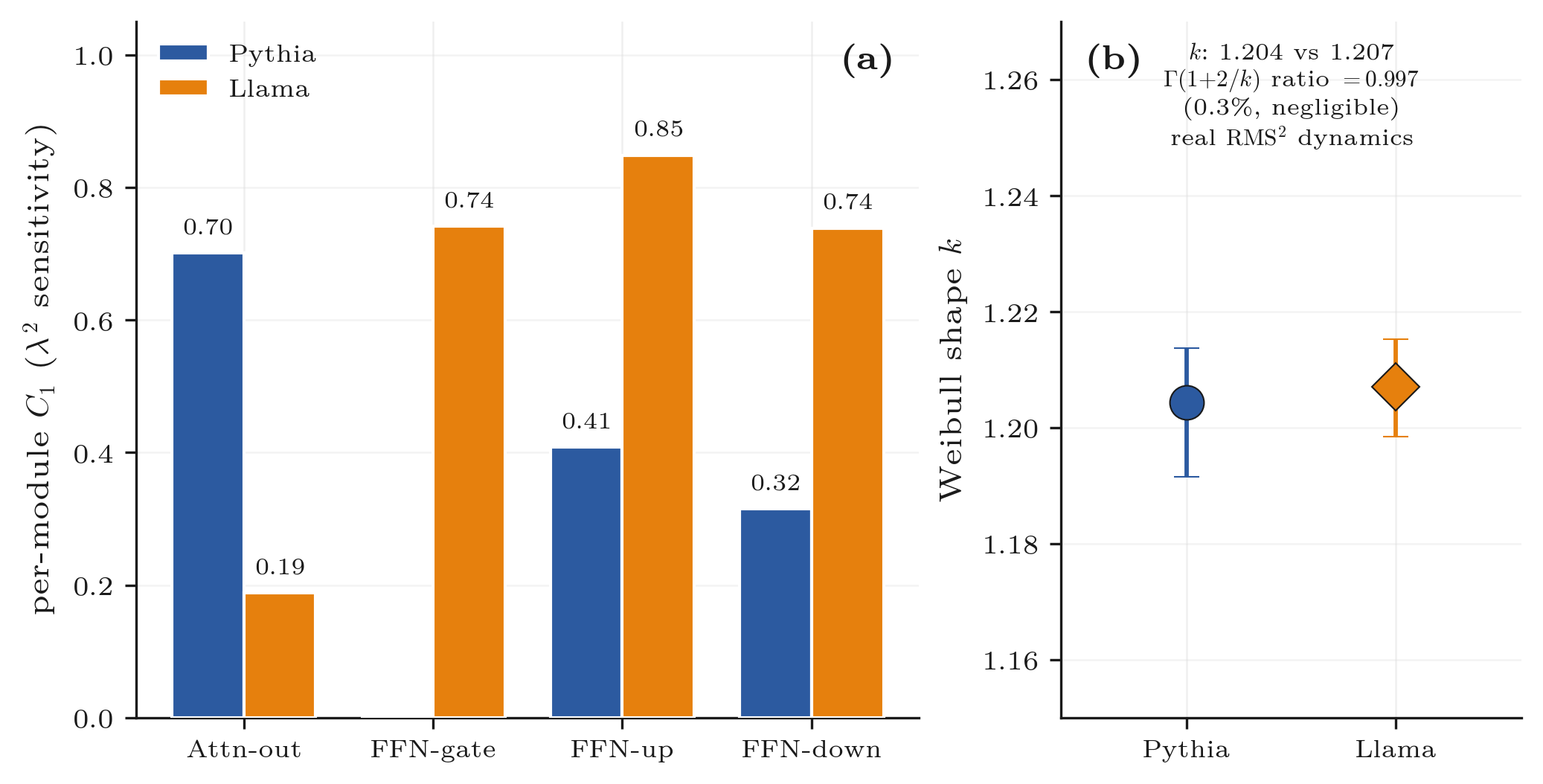}
  \caption{\textbf{The architecture difference is in the weights, not the ruler.} (a) per-module data
    sensitivity $\Cone$ in the common $\lambda^2$ currency: Llama's SwiGLU feed-forward
    (gate/up/down, $0.74$--$0.85$) is about $2\times$ Pythia's GELU feed-forward ($0.32$--$0.41$),
    while Pythia carries its sensitivity in the attention-output projection ($0.70$). FFN-gate has no
    Pythia counterpart (gating is SwiGLU-only), hence the single Llama bar. (b) the Weibull shape is
    matched ($k=1.204$ vs $1.207$; $\Gamma(1{+}2/k)$ ratio $0.997$), so the difference is consistent
    with an architecture-dependent $\lambda^2$ response, not a shape artifact. Single seed per architecture.}
  \label{fig:arch-mech}
\end{figure}

We first check that the difference is a property of the weights, not of the Weibull ruler. The fitted shape parameters are nearly matched --- median $k = 1.204$ (Pythia) vs $1.207$ (Llama), both near $k \approx 1.20$ --- so the second-moment conversion $\langle w^2\rangle = \lambda^2\,\Gamma(1+2/k)$ differs by a $\Gamma$-ratio of only $0.997$ ($0.3\%$; \cref{fig:arch-mech}, right), and the two architectures are read on essentially the same scale. Across all fourteen runs the $\lambda$ trajectories rise monotonically without overshoot, and the gated architecture's per-step alignment forces are no larger than the ungated one's. The steeper Llama slope is therefore an architecture-dependent weight-scale ($\lambda^2$) response, not a by-product of a different distribution shape or of a SwiGLU loss-spike amplified through the Adam update.

The localization points to the feed-forward pathway. Decomposing $\Cone$ per module, in the additive $\lambda^2$ currency of the layer analysis below (\cref{fig:arch-mech}, left; \cref{fig:arch-perlayer}), the Llama-style SwiGLU feed-forward blocks (gate / up / down, $\Cone \approx 0.74$--$0.85$) are about $2\times$ as data-sensitive as Pythia's GELU feed-forward ($0.32$--$0.41$), and respond at every depth, whereas the GELU feed-forward turns sensitive only in the deeper blocks; Pythia instead carries more of its response in the attention-output projection (per-$\rho$ $\lambda$ maps, \cref{fig:arch-lam-maps}; end-of-training depth profile, \cref{fig:arch-perlayer-lam}). Architecture thus enters the law through its response coefficients --- a gated nonlinearity converting predictable structure into weight-scale growth more efficiently than an additive one --- while the convex form and its data-side exponent $0.59$ are unchanged. Isolating the gating from the normalization and residual-scaling differences that also distinguish the two would require a controlled single-factor swap, which we leave to future work (\cref{sec:discussion}).

\subsection{Layers, and the relation between pooled and per-layer readouts}
The same Weibull ruler reads at two granularities: a pooled \emph{model-level} $(k,\lambda)$ that summarizes how the whole model responds to the corpus, and a \emph{block-level} $(k,\lambda)$ per block that localizes where that response is written. What connects the two is not $\lambda$ itself but the second moment. For a Weibull fit, $\langle w^2\rangle = \lambda^2\,\Gamma(1+2/k)$; because second moments add under pooling while $k$ stays nearly constant across blocks, the additive currency is $\lambda^2$ (up to the nearly constant $\Gamma$-factor). The scale $\lambda$ is \emph{not} additive and should never be summed or averaged directly. This is why the law is stated in $\dlamsq$, not in $\lambda$: second moments add; scales do not.

\begin{figure}[t]\centering
  \includegraphics[width=0.85\linewidth]{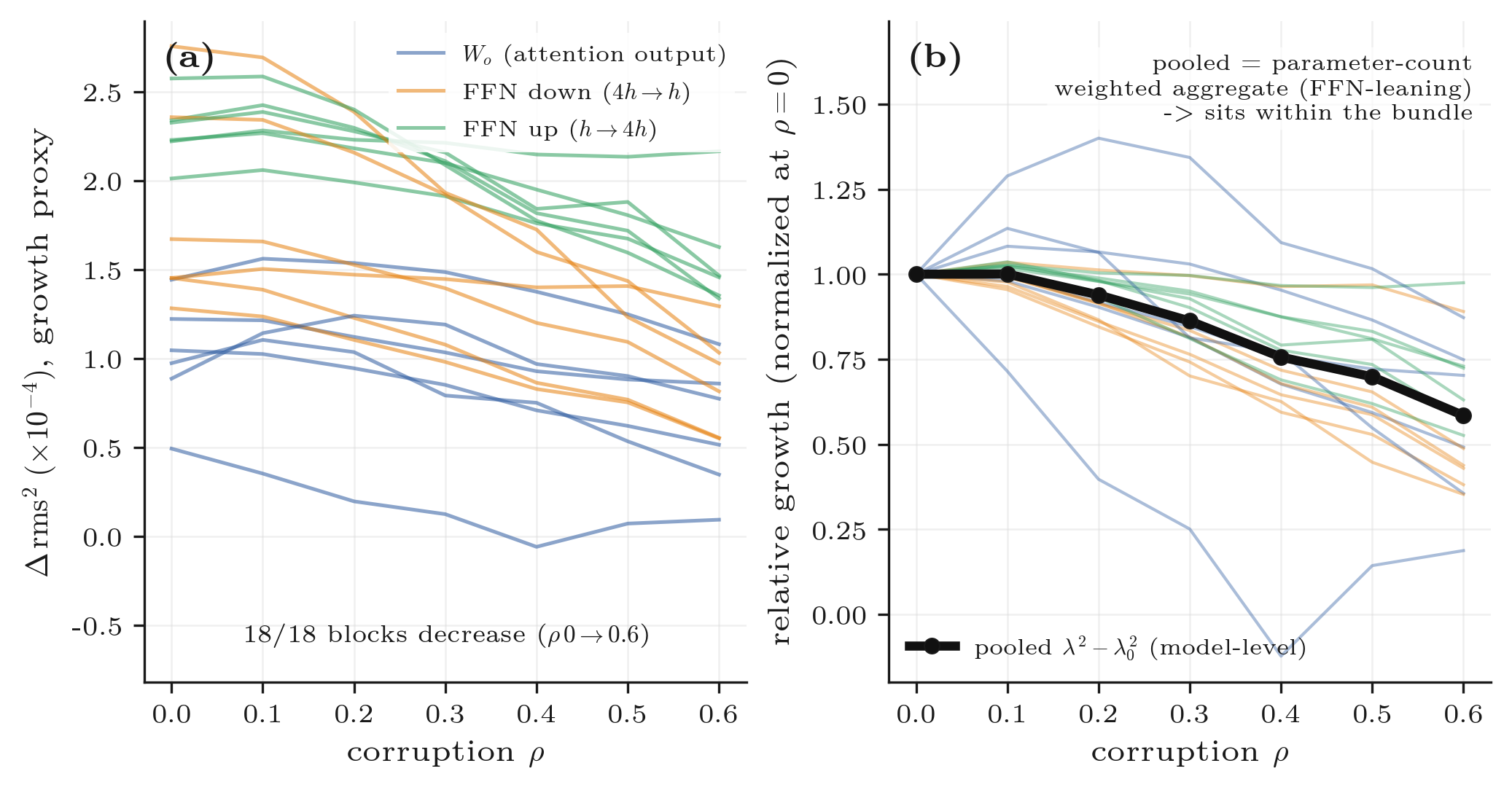}
  \caption{\textbf{The model-level law is not a pooling artifact.} (a) all 18 blocks (6 layers
    $\times$ 3 components) decrease as the data degrades, feed-forward most; (b) normalized, the
    pooled $\dlamsq$ (bold) sits within the per-block bundle and equals the parameter-count-weighted
    aggregate of the per-block laws. Shape $k$ uniform across blocks ($1.16$--$1.21$); the law holds
    per block (median $R^2=0.84$).}
  \label{fig:perlayer}
\end{figure}

This additivity is what makes the model-level law more than a pooling artifact. Fitting \cref{eq:law} block by block (\cref{fig:perlayer}, 18 blocks = 6 layers $\times$ 3 components), the same data-side regressor $\pmar$ applies to every block --- it is a property of the \emph{corpus} --- and only the response coefficients $\Czero,\Cone$ change. The blockwise fits keep the same functional form, with median $R^2 = 0.84$ (the cleanest feed-forward-down blocks reach $0.92$; early and attention-output blocks are noisier), and $k$ stays uniform across blocks ($1.16$--$1.21$), supporting a common second-moment currency.

We verify the aggregation directly. Pooling the per-block $\lambda^2$ by parameter count and \emph{then} fitting \cref{eq:law} gives the same coefficient as the parameter-weighted average of the per-block slopes $\Cone$, to numerical precision ($10^{-15}$); an unweighted average or a sum does not (Llama: $0.71$ weighted vs.\ $0.63$ unweighted vs.\ a meaningless $15.1$ summed), confirming that only the additive $\lambda^2$ currency may be aggregated, never $\lambda$ itself. Restricted to the instrumented transmission-class matrices, the weighted aggregate recovers $\Cone = 0.40$ (Pythia) and $0.71$ (Llama), close to the directly fitted model-level $0.45$ and $0.78$; the residual is the uninstrumented matrices (query/key/value projections, embeddings) and the mixture approximation (a mixture of same-$k$ Weibulls is only approximately Weibull), not a breakdown of the decomposition.

The two readouts therefore play complementary roles. The pooled model-level law is a fast \emph{screen} --- a single $(k,\lambda)$ pair per model, tied to the corpus statistic $\Dc$ --- that flags whether and how strongly a model's weight scale tracks data predictability. The per-block readout then \emph{localizes}, pointing to which blocks carry the response. Used together, they turn one model-level number into a map of where data predictability is written into the network.

\section{Limitations and Scope}\label{sec:limitations}

This section states the scope under which the law should be read. The controlled result is a within-corpus law for local mapping predictability; it is not yet a universal cross-corpus data-quality metric.

\begin{enumerate}
\item \textbf{$\Hr$ is an estimator ceiling, not the random entropy.} The anchor $\Hr = 8.0$ is the value the plug-in bigram conditional entropy saturates to at full shuffle under our 2.4M-token budget, not the information-theoretic marginal entropy of the corpus (10.7 bits): with $\sim\!10^{6}$ samples over a $\sim\!10^{9}$-cell bigram space, the plug-in estimate is biased low. $\Hr$ should be read as a matched-budget shuffle baseline --- the same saturation that makes higher-order or held-out entropy estimators fail to restore resolution at high $\rho$.

\item \textbf{The law is within-corpus.} $\pmar$ is not a universal cross-corpus currency. Code, whose $\Dc=5.30$ is lower than any corruption level, falls well below the law (\cref{fig:2dredund}) and is over-predicted by $\sim 6\times$ the in-corpus error (\cref{fig:selfval}, \cref{app:selfval}); mapping-rich corpora (wikitext, c4) are predicted well. We mark this boundary rather than fit through it --- its cause is the redundancy dimension noted next.

\item \textbf{Baseline renormalization does not remove the code failure.} A per-corpus shuffle ceiling $\Hr^{\text{corpus}}$ does not make the margin comparable across corpora: the plug-in shuffle ceiling is $\approx 8.0$ for every corpus tested (it is set by the token budget, not the corpus), so self-normalization still leaves code's margin \emph{larger} than wikitext's even though code grows \emph{less}; the true-marginal version fails the same way. The failure is not a baseline-choice artifact.

\begin{figure}[H]\centering
  \includegraphics[width=0.80\linewidth]{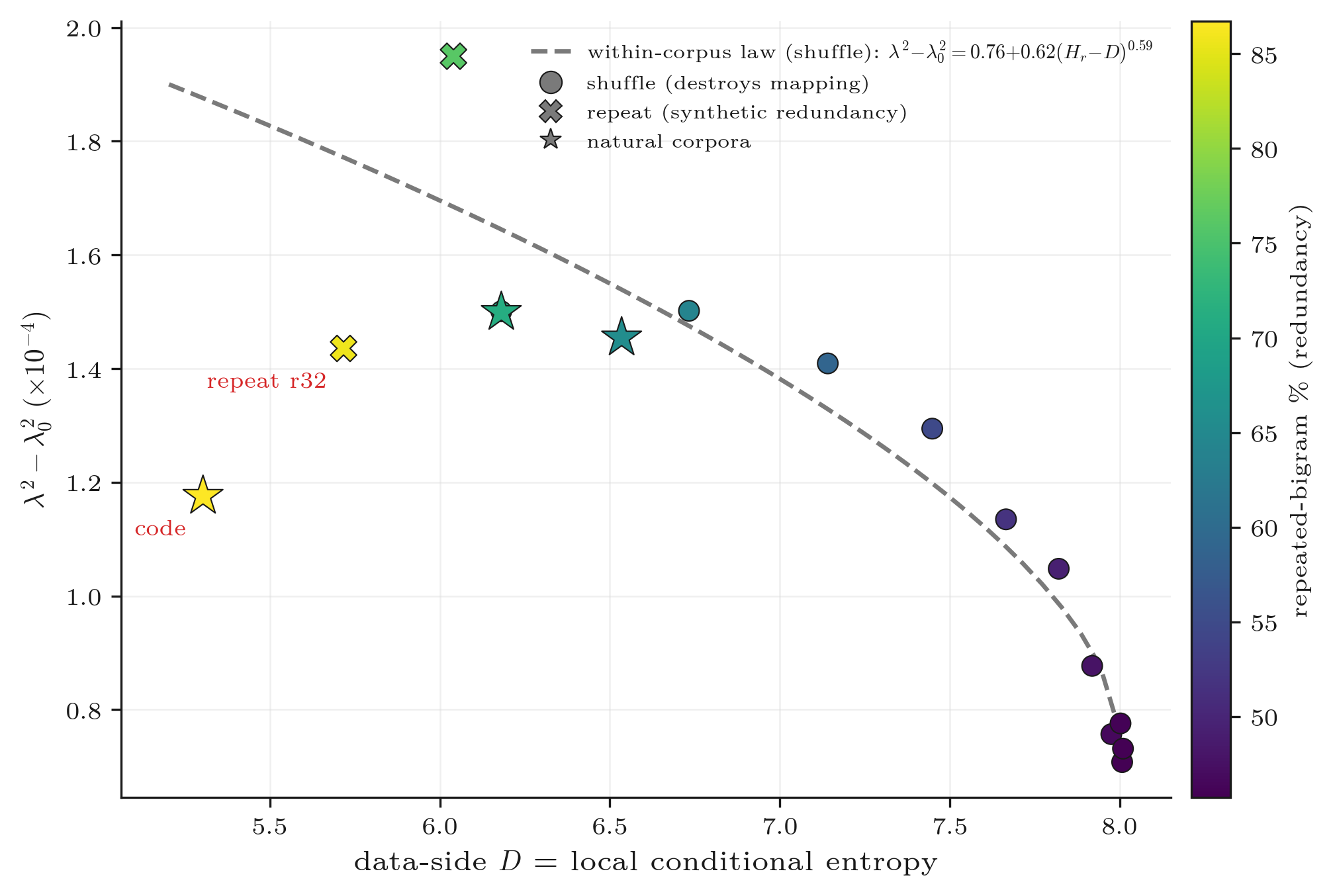}
  \caption{\textbf{A second, redundancy axis.} Growth vs $\Dc$, points coloured by repeated-bigram
    fraction (redundancy). The corruption gradient (wikitext shuffle, $\rho=0\to1$) defines the
    within-corpus law (dashed), which mapping-rich wikitext and c4 follow; high-redundancy data
    --- code and synthetically repeated text --- falls below it. This implicates redundancy as the
    most parsimonious second axis consistent with code's departure (developed in \cref{sec:discussion}); the
    figure is illustrative, not a fitted two-dimensional law.}
  \label{fig:2dredund}
\end{figure}

\item \textbf{Redundancy is the likely missing data dimension.} A single conditional-entropy variable conflates rich-mapping-low-$\Dc$ with redundant-low-$\Dc$: code and $32\times$-repeated text register as equally redundant under $\Dc$, yet train differently. Code's $\Dc$ is also the least stable across disjoint subsamples (standard error $\approx 0.094$ bits, an order of magnitude above the corruption corpora at $\approx 0.003$--$0.011$), consistent with templated, heterogeneous statistics rather than a single homogeneous mapping. This implies $\dlamsq$ is at least a two-variable function of mapping structure and redundancy. The evidence rests on a single off-law corpus, so redundancy is the most parsimonious second axis rather than an identified one; we develop this $(D{+}R)$ direction --- and the corpus design that would decouple the two axes --- in \cref{sec:discussion}.

\item \textbf{Coefficients are architecture-specific.} The data-side \emph{form} is shared across architectures, but the coefficients $\Czero(\eta), \Cone(\eta)$ are not (\cref{sec:robustness}): a $\lambda$-based data diagnostic calibrated on one model family does not carry its coefficients to another, so cross-family use requires per-architecture recalibration. We establish this only at the form level --- two families, one seed each, with the feed-forward gate confounded by the normalization (RMSNorm vs LayerNorm) --- so the \emph{magnitude} of the cross-architecture difference is indicative, and isolating its cause needs the controlled ablation deferred to \cref{sec:discussion}.

\item \textbf{Model and training scale.} The law is established at a single model scale ($\sim$70M parameters) and a single training budget ($\approx 10^{8}$ tokens over 8000 steps), well inside the transient regime --- far short of the weight-decay steady state $\tau\approx 1/(\eta\lambda_{\text{wd}})$ (\cref{fig:law-step}). The corruption corpora are also derived from a single base corpus, and the natural-corpus boundary rests on a few held-out corpora. Whether the convex form and its coefficients carry to substantially larger models, longer near-steady-state training, or a broader corpus population is untested here; we read the result as a controlled small-scale law and leave its scale-dependence to future work.

\item \textbf{Classical metrics; estimator saturation.} The statistics are classical --- $\Dc$ is a conditional-entropy cousin of perplexity, the alignment quantity a weight--gradient cosine --- and we claim the closed loop and the derived form, not the metrics (\cref{sec:related}). The plug-in bigram estimator saturates above $\rho \approx 0.6$, which is why the main law is read in the non-saturated range.

\item \textbf{Replication and causal scope.} Some $\rho$ levels and learning rates are single-seed, so their coefficient magnitudes are indicative rather than final. Correlations along the corruption axis are not claims of causal mediation; what the paper establishes is a controlled predictive relation from a pre-training data statistic to a post-training weight-scale outcome.
\end{enumerate}

\section{Related Work}\label{sec:related}

\textbf{Gradient--weight alignment.} Gradient--weight alignment provides a natural precedent for using weight-update coherence as a mechanistic quantity: Gradient--Weight Alignment \citep{holzl2025gwa} uses the cosine between per-sample gradients and model weights to quantify gradient--weight coherence, for generalization and single-sample attribution in supervised image classification. We use an aggregate alignment channel for a different purpose --- connecting corpus-level predictability to Weibull weight-scale growth through a controlled corpus swap.

\textbf{Data complexity and the stages of training.} A line of work maps \emph{data/function complexity} --- statistical-moment order, token $n$-gram structure, many-body interaction order --- onto the stages of learning (the distributional simplicity bias) \citep{refinetti2023sgd, belrose2024statistics, rende2024distributional}. Our work is complementary: rather than assigning complexity stages in function space, we recover a data-dependent training outcome from the weight trajectory and express it as a scale-growth law --- the model-conditioned (weight-trajectory) counterpart of their model-agnostic (data-complexity) view.

\textbf{Weight-magnitude and spectral data diagnostics.} Several works read data properties from the weights' heavy-tailed \emph{spectrum}: the empirical spectral density tail index detects label noise \citep{loftus2026spectral}, dataset diversity shapes the weight spectrum \citep{ba2024diversity}, and classification difficulty drives a heavy-tail phase transition \citep{meng2023impact}. These differ from our readout on three axes: a spectral tail exponent $\alpha$ versus the entry-wise Weibull scale $\lambda$ (with $k$ fixed); a static snapshot versus the training trajectory; and small vision/classification models versus a language-model corpus statistic. Reading data dependence from the \emph{dynamics} of the weight spectrum is closest in spirit to \citep{yunis2024spectral}, which however tracks generalization rather than a pre-training data statistic. That this line reports norm and spectral measures breaking under hyper-parameter changes is consistent with our learning-rate-conditioned response, which we model through $\Czero(\eta), \Cone(\eta)$ (\cref{sec:law}).

\textbf{Geometry versus data in the scale.} AdamW rotational-equilibrium analyses explain how optimizer dynamics shape weight norms and scales \citep{kosson2024rotational, chou2025correction}. Our question is complementary: how the \emph{data-dependent} component of that dynamics appears in Weibull $\lambda$-growth. In this view, optimizer geometry sets the scale dynamics, while corpus structure modulates the alignment channel and hence the fitted response coefficients $\Czero(\eta), \Cone(\eta)$.

\textbf{Controlled data interventions.} Controlled data interventions have been used to isolate effects on downstream behavior --- knowledge, memorization, accuracy \citep{bordt2026trainonce}. Our intervention instead holds the training setup fixed, swaps the corpus under a fixed base model, and reads the response directly in the weight distribution.

\textbf{Architecture as a modulator of the data-to-weight map.} That the same law holds across architectures with only its coefficients moving (\cref{sec:robustness}) draws together four neighboring lines. \emph{Loss-side scaling:} loss-to-loss prediction finds power laws between the losses of models trained on different datasets \citep{brandfonbrener2024losstoloss} --- the loss-side analogue, explicitly not about weights. \emph{Representation-side layered learning:} that shallow layers capture surface/token statistics and deep layers task-specific structure is established in representation space \citep{tenney2019bert} and the distributional-simplicity-bias line above; our depth result is its weight-side counterpart, a per-layer weight-\emph{scale} response to data corruption. \emph{Update scaling:} the maximal-update / Tensor Programs line governs how per-layer update magnitudes scale with width and depth for stable feature learning \citep{yang2022tensorprograms}, for stability rather than for how data predictability is written into weights. \emph{Information in weights:} the nearest conceptual predecessor \citep{achille2018emergence} quantifies the mutual information a trained weight carries about the data, but through an information bottleneck rather than a magnitude-distribution scale and with no architecture or data-predictability axis. The gated-feed-forward reading we invoke rests on the multiplicative gradient structure of GLU variants \citep{dauphin2017gated, shazeer2020glu}, whose effectiveness their own authors leave open; we state it as a hypothesis. We are not aware of prior work that combines these ingredients into a weight-scale law in which data predictability is the independent variable and architecture modulates the response coefficients.

\textbf{Summary.} Taken together, prior work provides several neighboring ingredients --- alignment quantities, data-complexity measures, spectral weight diagnostics, optimizer-scale geometry, and controlled data interventions. The present work connects them in a single weight-side law: a pre-training data statistic predicts Weibull weight-scale growth, learning rate and architecture enter through response coefficients, and the same readout serves at both the model and layer levels.

\section{Discussion and Future Work}\label{sec:discussion}

\textbf{A response surface over data, architecture, and optimization.} It helps to read these results as one slice of a larger object: a response surface $\Phi(\Dc, R, A, H)$ that maps two data-side axes --- mapping predictability $\Dc$ and redundancy $R$ --- together with the architecture $A$ and the optimization hyperparameters $H$ (learning rate, budget; $H$ denotes hyperparameters, not the entropy ceiling $\Hr$) to Weibull weight-scale growth. This paper establishes the $\Dc$ axis. Because $\Dc$ is computed before training, the law is a forward predictor of weight-scale growth from data, and within a corpus the prediction is quantitative (\cref{app:selfval}). It is also bounded: $\pmar$ is not a cross-corpus currency (\cref{sec:limitations}), so general data-quality diagnosis would need the remaining axes and a reference-model perplexity proxy to carry $\Dc$ to natural corpora. We read $\Phi$ axis by axis below --- what we have fixed, what we have begun to vary, and what stays open.

\textbf{The redundancy axis $R$: repetition and multi-epoch training.} Holding architecture and optimization fixed, we vary a second data-side axis. Tiling a fraction of the corpus is, at a fixed token budget, equivalent to training several epochs on that fraction. Weight-scale growth is non-monotonic in the repetition factor (\cref{fig:repeat}): it rises from the single-pass baseline, peaks near an eight-fold repeat, then falls back below baseline by thirty-two-fold (single seed). This inverted-U sits alongside the diminishing returns of repeated data documented for data-constrained training by \citet{muennighoff2023scaling}: where they measure a loss-side saturation of repeated tokens, we observe its weight-side counterpart in $\lambda$, with the turnover at a comparable scale of a few epochs. A possible reason is a balance of two effects --- moderate repetition reinforces the alignable structure already present, while excessive repetition exhausts the unique structure available to align to --- which is consistent with the three-force account, though we do not establish it as the mechanism. As a single-seed lead outside the one-dimensional law this is deferred to future work; it does suggest $\lambda$ as a candidate weight-side diagnostic of over-repetition.

\textbf{The architecture and optimization axes $A$ and $H$: coefficients move, the form does not.} \Cref{sec:robustness} is a first cut at these axes. The hyperparameters enter only through the coefficients, which follow power laws in the learning rate (\cref{fig:eta}). Across Pythia and Llama the picture is the same in kind: the data law keeps its sign, its convex shape, and its data-side exponent, and only the coefficients $\Czero(\eta), \Cone(\eta)$ move (\cref{sec:robustness}). Read as a metric, $\Cone$ is a weight-side \emph{data-to-weight conversion efficiency}: of the predictable structure a corpus offers, how much each block writes into its weight scale. Two correlates are visible in the per-block maps (\cref{fig:arch-perlayer,fig:arch-lam-maps}) and we report them as observations, not mechanism. First, the feed-forward nonlinearity: Llama's gated SwiGLU is about twice as data-sensitive as Pythia's GELU, and at every depth rather than only the deep blocks. Second, depth: in both models the shallowest blocks are the least sensitive --- near zero, and formally inverse at the first block --- and sensitivity rises with depth. We estimate that the gate and this depth ordering are what these differences track, but with one seed per architecture and a black-box A-vs-B comparison --- the two families also differ in normalization, a known control on steady-state weight scale \citep{kosson2024rotational} --- we cannot separate the gate from these confounds here; a controlled single-factor ablation is left to future work (\cref{sec:limitations}).

\textbf{Toward the full surface.} The two data axes we have handles on --- mapping $\Dc$ and redundancy $R$ --- appear together in \cref{fig:2dredund}: corpora that share a low $\Dc$ but differ in redundancy (corrupted text versus repeated or templated content such as code) separate in the plane, and the off-law corpora fall away from the one-dimensional fit along the $R$ direction. This is the $(\Dc, R)$ face of $\Phi$, and it is what motivates treating redundancy as an axis in its own right rather than as noise on $\Dc$. We expect data to carry further independent axes, each with its own pre-training measure and its own training-dynamics signal --- a semantic axis (embedding similarity $\leftrightarrow$ gradient signal-to-noise / $k$) and a directional axis among them. Mapping the full $\Phi$ surface over these axes, with architecture and optimization as further arguments, is the program these results open.

\textbf{Summary.} Three threads run through these results: data predictability is not one number but several measurable axes; the architecture enters as a layer of coefficients on a fixed law; and one pair of Weibull parameters supports both global diagnosis and per-layer localization of where that predictability is written. Throughout we have held to two disciplines --- establish one data axis thoroughly rather than many partially, and require the loop to close in both directions, from a data-side measurable to a weight-side quantity and back.

\section{Conclusion}\label{sec:conclusion}

We establish a within-corpus, learning-rate-conditioned law linking local next-token predictability in the training data to the growth of the Weibull weight-scale parameter. Its convex form is explained by the saturation of bigram entropy against retained local structure, and the law is validated end-to-end by predicting held-out within-family weight growth from a pre-training data statistic. The same readout works at both model and per-layer scales, connecting global diagnosis with layer-level localization.

The boundary is equally informative: the law predicts mapping-rich text but fails on code, implicating redundancy as a second data dimension rather than a failure of the Weibull readout itself. Thus the result should not be read as a universal data-quality metric. It is a controlled bridge from a measurable corpus statistic to weight-scale dynamics, and it provides the first calibrated axis of a broader $\Phi(\Dc, R, A, H)$ framework in which data structure, redundancy, architecture, and hyperparameters can be studied one dimension at a time.

\textbf{Code and data.} The training and corruption scripts and the fitted $\lambda$ trajectories are released at \url{https://github.com/tiexinding/NPM-Weibull-public}, within the unified repository for this line of work; the corruption corpora are reconstructed from the public sources cited in \cref{sec:method}.

\bibliographystyle{tmlr}
\bibliography{references}

\appendix

\section{Notation}\label{app:notation}
The symbols used throughout are summarized in \cref{tab:notation}. The convex-law derivation additionally classifies each quantity as measured, fitted, or derived in \cref{tab:B1}.

\begin{table}[H]
\centering
\caption{Summary of notation.}
\label{tab:notation}
\small
\begin{tabular}{p{2.7cm}p{11.6cm}}
\toprule
symbol & meaning \\
\midrule
\multicolumn{2}{l}{\emph{Weibull weight readout}}\\
$\lambda$ & Weibull scale parameter of a weight-magnitude distribution (the quantity that moves during training) \\
$k$ & Weibull shape parameter ($\approx1.20$, stable across blocks and models; reserved for the shape, never the law slope) \\
$\lamz$ & initial scale, at the first checkpoint; subtracted to isolate growth \\
$\dlamsq$ & weight-scale growth --- the \emph{additive} second-moment currency in which the law is stated \\
$\Gamma(1+2/k)$ & second-moment conversion factor, $\langle w^2\rangle=\lambda^2\,\Gamma(1+2/k)$ \\
\midrule
\multicolumn{2}{l}{\emph{Data side (computed from the corpus before training)}}\\
$\Dc$ & bigram conditional entropy $H(\mathrm{next}\mid\mathrm{prev})$; a training-free measure of local next-token predictability \\
$\Hr$ & matched-budget plug-in shuffle ceiling of $\Dc$ ($\approx8.0$); \emph{not} the information-theoretic random entropy ($10.7$ bits) \\
$\pmar$ & predictability margin --- the law's data-side variable \\
$\rho$ & corruption level (fraction of within-sequence positions shuffled); $0$ clean, $1$ scrambled \\
$S$ & struct-retain fraction (adjacent token pairs preserved vs.\ the clean corpus); runs $1\!\to\!0$ with corruption; experiment-internal \\
\midrule
\multicolumn{2}{l}{\emph{The law}}\\
$\eta$ & learning rate \\
$\Czero(\eta)$ & injection floor (intercept): the growth surviving at zero predictive structure \\
$\Cone(\eta)$ & alignment efficiency (slope): the data-to-weight conversion efficiency \\
$p$ & data-side saturation exponent ($\approx1.69$; $\Dc=\Hr-\Delta H\,S^{p}$) \\
$0.59$ & convex-law exponent, $=1/p$ \\
$\Delta H$ & predictive-information range, $=\Hr-D_{\mathrm{clean}}\approx1.82$ \\
\midrule
\multicolumn{2}{l}{\emph{Response surface (\cref{sec:discussion})}}\\
$\Phi(\Dc,R,A,H)$ & response surface mapping the two data axes, the architecture, and the optimization onto weight-scale growth; this paper traces the $\Dc$ axis and gives first cuts on the others \\
$R$ & redundancy axis: fraction of repeated / templated content --- a second data-side axis beyond $\Dc$ \\
$A$ & architecture: model family and feed-forward nonlinearity \\
$H$ & optimization hyperparameters (learning rate, training budget); denotes hyperparameters, \emph{not} the entropy ceiling $\Hr$ \\
\midrule
\multicolumn{2}{l}{\emph{Derivation-internal (appendix)}}\\
$a_0,a_1$ & intercept/slope of the training-side fit in $S$ (\cref{eq:train-linear}); the law's $\Czero,\Cone$ in the $S$ coordinate ($\Czero{=}a_0$, $\Cone{=}a_1/\Delta H^{0.59}$) \\
\bottomrule
\end{tabular}
\end{table}

\section{\texorpdfstring{End-to-End Self-Validation: Predicting $\lambda$ from Data Alone}{End-to-End Self-Validation: Predicting lambda from Data Alone}}\label{app:selfval}

\subsection{Logic}

The law (\cref{eq:law}/\cref{sec:convex}) takes a quantity computed from the \emph{raw corpus before any training} (the bigram conditional entropy $\Dc$) and predicts the weight-scale growth $\dlamsq$. We can therefore test it as a genuine forward prediction: compute $\Dc$ from a corpus, predict $\dlamsq$ from the data alone, then actually train and measure $\dlamsq$ from the Weibull fit of the weights. The two are the \emph{same quantity} obtained two independent ways, so they are directly comparable. The strength of the test is set by one condition: \textbf{the predicted data does not enter the coefficient fit.}

We report three findings: (i) the data-side rung of the derivation is near-exact; (ii) on held-out corruption levels of the same corpus the chain predicts $\lambda$ to 5.7\% relative error; and (iii) on a held-out \emph{corpus} the chain succeeds for mapping-rich text and fails --- as predicted --- for code, quantifying the boundary of \cref{sec:limitations}.

\subsection{Protocol}

The prediction chain (\cref{sec:convex}) is $\Dc \to S \to \lambda^2$. We exploit that the struct-retain fraction $S$ is \emph{independently measurable} from the corruption recipe (not only invertible from $\Dc$), which lets us test the two rungs separately:

\begin{itemize}
\item \textbf{B rung (data side, fixed):} $\Dc = \Hr - \Delta H\,S^{1.69}$ with $\Hr=8.0,\ \Delta H=1.82$ fixed from history. We test it by predicting $\Dc$ from the measured $S$.
\item \textbf{A rung (training side, fitted):} $\dlamsq = a_0 + a_1 S$, with $a_0, a_1$ fitted only on training data.
\item \textbf{Full chain (data-only prediction):} invert B to get $S=((\Hr-\Dc)/\Delta H)^{1/1.69}$ from the measured $\Dc$, then apply A. This uses \emph{no} training-side information about the predicted run. \emph{Note:} this composition is also what ties $a_0, a_1$ to the main law --- substituting the inverted B into A gives $\dlamsq = a_0 + (a_1/\Delta H^{0.59})\,\pmar^{0.59}$, so $a_0, a_1$ are not a separate pair but the law's $\Czero, \Cone$ in the $S$ coordinate: $\Czero = a_0$ and $\Cone = a_1/\Delta H^{0.59}$, with the data-side factor $\Delta H^{0.59}$ the same for every $\eta$ (\cref{app:construction}). The $a$-symbols are kept because the derivation measures the training side as linear in the experiment-internal $S$; the main $\eta$-conditioned law (\cref{eq:law}) is stated in $\Czero, \Cone$.
\end{itemize}

\textbf{L1 --- held-out corruption levels (leave-one-out).} Same corpus (wiki), same architecture (Pythia-70m), $\eta=3\text{e-}4$, realistic regime $\rho \le 0.6$ (7 levels; $\rho \in \{0,0.2,0.4,0.6\}$ seed-averaged, $n=3$). For each level we fit the A rung (the training-side $S\to\lambda^2$ relation) on the other six and predict the held-out one.

\textbf{L2 --- held-out corpus.} Fit the A rung on the realistic-regime ($\rho\le0.6$) wiki corruption family (the same regime as L1), then predict three natural corpora from their measured $\Dc$ alone. The coefficients are fit \emph{once} on the wiki corruption family and then frozen: the measured $\Dc$ of each natural corpus is passed through the same B inversion and A-rung predictor, and no training outcome from c4 or code is used to fit the predictor. Because the corruption family is built from wikitext, its clean point already enters the coefficient fit: wikitext is therefore an \textbf{in-family reference} --- a consistency check, not a blind natural-corpus prediction. The genuine held-out tests are \textbf{c4} (a mapping-rich natural text) and \textbf{code} (a redundancy-dominated corpus), each $n=3$ seeds, neither entering the fit.

\subsection{Results}

\textbf{B rung: the data-side inversion adds negligible error.} Predicting $\Dc$ from the measured $S$ gives $R^2 = 0.9997$, RMSE $= 0.011$ bits. We do not read this as an independent discovery --- $S$ and $\Dc$ are two measurements of the same corrupted data, and the saturation relation between them is the data-side fit of this corruption family --- but as confirming that the data-side rung is stable out-of-sample, so that whatever residual the chain carries comes from the A rung, not the inversion.

\textbf{L1 (\cref{tab:A1}, \cref{fig:selfval} panel a).} The observed $\dlamsq$ values are measured after training, from Weibull fits to the transmission-class weight matrices, using the same readout as the main text. Leave-one-out blind prediction gives RMSE $= 0.092$ ($\times 10^{-4}$), a \textbf{5.7\% relative error}, versus an in-sample RMSE of 0.054 (overfitting ratio $1.71\times$). The full-chain error equals the A-rung error to three decimals (\cref{fig:selfval} panel b): the B inversion adds nothing, so the residual is entirely the linear-in-$S$ scatter of the A rung, largest at the two ends ($\rho=0$ and $\rho=0.6$) where the convex curvature is strongest --- the same curvature derived in \cref{sec:convex}. The 5.7\% is an RMSE aggregate; the worst single-point error is at $\rho=0$ (the in-family reference), where the linear-in-$S$ A-rung over-predicts the clean end by 0.189 ($\approx 12.6\%$) --- where the convex bend the linear model omits is strongest. We report both the aggregate and this worst case.

\begin{table}[t]
\centering
\caption{L1 leave-one-out blind prediction ($\eta=3\text{e-}4$, $\rho\le 0.6$; all $\dlamsq$ values are $\times 10^{-4}$). The relative error $|\text{obs}-\text{pred}|/\text{obs}$ matches the full-chain bars of \cref{fig:selfval}(b). Here obs denotes the measured post-training Weibull $\lambda^2$ growth.}
\label{tab:A1}
\begin{tabular}{lccccccc}
\toprule
$\rho$ & $n_{\text{seed}}$ & $\Dc$ & $S$ & $\dlamsq$ obs & pred (LOO) & err & rel.\ err (\%) \\
\midrule
0.0 & 3 & 6.18 & 1.000 & 1.503 & 1.692 & $-0.189$ & 12.6 \\
0.1 & 1 & 6.73 & 0.816 & 1.502 & 1.438 & $+0.064$ & 4.3 \\
0.2 & 3 & 7.14 & 0.647 & 1.397 & 1.325 & $+0.071$ & 5.1 \\
0.3 & 1 & 7.45 & 0.496 & 1.294 & 1.221 & $+0.073$ & 5.6 \\
0.4 & 3 & 7.67 & 0.366 & 1.147 & 1.140 & $+0.007$ & 0.6 \\
0.5 & 1 & 7.82 & 0.257 & 1.048 & 1.068 & $-0.020$ & 1.9 \\
0.6 & 3 & 7.92 & 0.167 & 0.940 & 1.033 & $-0.093$ & 9.9 \\
\midrule
\multicolumn{7}{r}{mean} & 5.7 \\
\bottomrule
\end{tabular}
\end{table}

\textbf{L2 (\cref{tab:A2}, \cref{fig:selfval} panel a, stars).} With coefficients fit only on the realistic-regime wiki corruption family ($a_0=0.88,\ a_1=0.71$ over $\rho\le0.6$, distinct from the full-range derivation coefficients $a_0{=}0.75, a_1{=}0.90$ of \cref{app:construction} and \cref{fig:deriv}a), the chain matches the mapping-rich corpora --- the in-family reference wikitext (a consistency check, already in the fit) within $-0.12$ and the genuinely held-out c4 within $-0.08$ --- but \textbf{over-predicts code by 0.50 in magnitude} ($\text{obs}-\text{pred} = -0.499$, $\approx 6\times$ the L1 LOO RMSE). Code's low $\Dc=5.30$ is read by the law as ``highly predictable $\to$ large growth,'' but code's low entropy comes from syntactic redundancy rather than a rich next-token mapping, so it grows far less. This makes the \cref{sec:limitations} boundary quantitative: the predictor performs well in the mapping-rich regime and fails in the redundancy-dominated regime anticipated by \cref{sec:limitations}, where a second (redundancy) dimension takes over.

\begin{table}[t]
\centering
\caption{L2 cross-corpus blind prediction using wiki-family coefficients. The A rung is fit only on the realistic-regime wiki corruption family, and no coefficients are refit on c4 or code. obs denotes measured post-training Weibull $\lambda^2$ growth; pred is computed from the corpus $\Dc$ alone through the fixed $\Dc\to S\to\lambda^2$ chain.}
\label{tab:A2}
\begin{tabular}{lccccc}
\toprule
corpus & $n$ & $\Dc$ & obs & pred & err (obs$-$pred) \\
\midrule
wikitext & 3 & 6.18 & 1.471 & 1.591 & $-0.120$ (in-family reference) \\
c4 & 3 & 6.53 & 1.426 & 1.506 & $-0.080$ (held-out) \\
\textbf{code} & 3 & 5.30 & 1.278 & 1.777 & $\mathbf{-0.499}$ (off-law, redundancy) \\
\bottomrule
\end{tabular}
\end{table}

\subsection{What this does and does not establish}

The self-validation is \emph{internal}. It shows that the proposed chain is self-consistent and predictive within the tested setting, not that it is the unique correct model. L1 should therefore be read as an internal predictive check rather than a formal pass/fail test. The aggregate LOO error is 5.7\%, with the worst single-point error 12.6\% at the clean endpoint, where the one-dimensional $S$-based A rung bends most strongly. Given that each held-out corruption level is predicted without using its own training outcome, this is a meaningful level of internal accuracy, but not a claim of exact prediction.

The stronger scope test is L2. The same wiki-family coefficients predict mapping-rich text reasonably well, including the in-family wikitext reference and the held-out c4 corpus, but over-predict code by a large margin. This failure is informative rather than incidental: code has low $\Dc$, but its low entropy is dominated by redundancy and templated structure rather than rich local next-token mappings. Thus L2 supports the boundary stated in \cref{sec:limitations}: the $\Dc$-axis law works where $\Dc$ measures mapping predictability, and breaks where redundancy becomes a separate data dimension.

The caveats are unchanged: one base corruption family, one main architecture, the realistic regime $\rho \le 0.6$, and limited seed coverage for some settings. The accompanying code and data tables make the prediction chain reproducible: $\Dc$, $S$, the A-rung fit, the B-rung inversion, and the final $\lambda^2$ predictions can be checked separately.

\section{\texorpdfstring{The $\rho$--$S$--$\Dc$ Construction and the Convex-Law Derivation}{The rho-S-D Construction and the Convex-Law Derivation}}\label{app:construction}

This appendix records the construction behind \cref{sec:convex}: how the corruption level $\rho$ produces the retained-structure variable $S$ and the bigram entropy $\Dc$, how the ceiling $\Hr$ and saturation exponent $p$ are obtained, and how eliminating $S$ gives the convex law in $\pmar$.

\subsection{One corruption knob, two measured consequences}

The corruption has a single input $\rho$; $S$ and $\Dc$ are two measured consequences of it, so they co-vary by construction. The relationship $\Dc=f(S)$ used in the derivation is therefore not the discovery of a link between two independent quantities --- it is the internal relationship between two images of the same knob, used to translate the training response from $S$-space into the measurable $\Dc$-space, after which $S$ is eliminated.

\textbf{$\rho \to S$ is essentially analytic.} Empirically $S \approx (1-\rho)^2$ to within $0.007$ across all eleven levels: relocating a fraction $\rho$ of tokens preserves an adjacent pair with probability $\approx(1-\rho)^2$ (neither endpoint moved). The small positive offset is the corpus's own coincidental-repeat rate (a displaced token occasionally lands an identical value), the first hint that even $S$ carries a trace of corpus statistics rather than $\rho$ alone. The map is strictly monotone, hence invertible.

\textbf{$\rho \to \Dc$ is monotone but saturates.} $\mathrm{d}\Dc/\mathrm{d}\rho$ falls from $5.5$ near $\rho=0$ to $0.02$ near $\rho=1$; the top three levels give $\Dc=8.002,8.007,8.010$, spacings ($0.005,0.002$) below the $\Dc$-estimation standard error ($\sim\!0.01$ bit). At the scrambled end $\Dc$ is therefore \emph{not} injective in $\rho$ --- several corruption levels are statistically indistinguishable in $\Dc$. This loss of resolution is the same saturation that produces the convexity and bounds the usable range to $\rho \lesssim 0.6$ (\cref{sec:limitations}).

\textbf{$S \to \Dc$ is monotone and near-deterministic, but not an identity.} The fit residual is $\le 0.014$ bit (rms $0.006$). $S$ counts adjacency; $\Dc$ measures bigram entropy; two shuffles with equal $S$ could in principle differ in $\Dc$. The tightness is an empirical property of \emph{this} corruption family, not a mathematical guarantee.

\begin{equation}
\begin{array}{c|ccccccc}
\rho & 0.0 & 0.2 & 0.4 & 0.6 & 0.8 & 0.9 & 1.0\\\hline
S & 1.000 & 0.647 & 0.366 & 0.167 & 0.044 & 0.012 & 0.000\\
\Dc & 6.181 & 7.141 & 7.665 & 7.919 & 8.002 & 8.007 & 8.010
\end{array}
\end{equation}

\subsection{\texorpdfstring{Determining $\Hr$ and $\Delta H$}{Determining H\_r and Delta-H}}

$\Hr$ is fixed two consistent ways: (i) as the free ceiling parameter of the B-step fit ($\Hr=8.008$), and (ii) directly as the measured $\Dc$ at full shuffle ($\rho=1$, $S\to0$: $\Dc=8.010$); the two agree to $0.002$ bit. It is a \emph{matched-budget plug-in shuffle baseline}, not the information-theoretic random entropy and not a universal constant (\cref{sec:limitations}).

$\Delta H$ is the fitted entropy range in the saturation model $\Dc = \Hr - \Delta H\,S^{p}$. It is close to $\Hr - \Dc^{\text{clean}}$ (here $8.008 - 6.181 \approx 1.827$) but is estimated jointly with $\Hr$ and $p$ rather than fixed exactly by the clean endpoint, giving $\Delta H \approx 1.82$. Within one corpus, corruption family, and token budget it is a fixed fitted constant; it can change with the corpus, the corruption mechanism, or the token budget.

\subsection{\texorpdfstring{From $a_0,a_1$ to $\Czero,\Cone$}{From a0,a1 to C0,C1}}

This is the single place that ties the training-side coefficients to the law's coefficients; the rest of the paper uses the relation without re-deriving it.

\textbf{(A) Training side --- linear in retained structure.} The alignment force is proportional to the learnable structure that survives, so
\begin{equation}
\dlamsq = a_0 + a_1\,S, \qquad a_0=0.749,\ a_1=0.896,\ R^2=0.945.
\end{equation}
The intercept $a_0$ is the injection floor (weights grow even at full shuffle); fitting through the origin gives $R^2=-1.9$, so the floor is real, not assumed.

\textbf{(B) Data side --- $\Dc$ saturates against $S$.}
\begin{equation}
\Dc = \Hr - \Delta H\,S^{p}, \qquad \Hr=8.008,\ \Delta H=1.818,\ p=1.691,\ R^2=0.9999.
\end{equation}
Here $p>1$, so $\mathrm{d}\Dc/\mathrm{d}S\to0$ as $S\to0$: $\Dc$ flattens at the scrambled end.

\textbf{(C) Composition --- eliminate $S$.} Solving (B) for $S=((\Hr-\Dc)/\Delta H)^{1/p}$ and substituting into (A),
\begin{equation}
\dlamsq = a_0 + a_1\Big(\tfrac{\Hr-\Dc}{\Delta H}\Big)^{1/p}
   = a_0 + \frac{a_1}{\Delta H^{1/p}}\,\pmar^{1/p},
\end{equation}
so $S$ cancels and the law is expressed in the measurable $\Dc$ alone, matching $\dlamsq=\Czero+\Cone\,\pmar^{\alpha}$ with
\begin{equation}
\Czero = a_0, \qquad \Cone = \frac{a_1}{\Delta H^{1/p}}, \qquad \alpha = \frac1p \approx 0.59 .
\end{equation}
At the reference $\eta=3\text{e-}4$, using the \cref{sec:convex} full-range derivation fit, this gives $\Czero=0.749$ and $\Cone = 0.896/1.818^{0.591} \approx 0.63$; the convex law fit directly over the same range agrees ($\Czero=0.751,\ \Cone=0.627,\ R^2=0.946$). These values hold under one condition --- the reference learning rate and the full-range derivation fit --- and should not be confused with the realistic-regime ($\rho\le0.6$) coefficients $a_0=0.88,\ a_1=0.71$ used for the cross-corpus test of \cref{app:selfval}. The conversion factor $\Delta H^{1/p}$ is a data-side quantity, the same for every $\eta$, so the $\eta$-dependence of $\Czero,\Cone$ lives entirely in $a_0,a_1$.

\subsection{Robustness of the saturation exponent}

The exponent is well-constrained for a stronger reason than the overall $R^2=0.9999$ (which co-monotone functions of $\rho$ could reach at many exponents). Fixing $p$ and refitting $(\Hr,\Delta H)$ gives a clear \emph{interior} optimum --- $R^2=0.954$ at $p=1.0$, $0.99988$ at $p=1.69$, $0.942$ at $p=3.0$ --- and leave-one-out over the eleven levels gives $p=1.688\pm0.014$. The reciprocal exponent is thus $0.59$ with a practical band of $\approx0.55\text{--}0.61$ (the upper end of $p$ coming from excluding the saturated plateau). This is what distinguishes the well-posed $S\!\to\!\Dc$ saturation fit ($S\in[0,1]$, $\Dc\in[6.18,8.01]$) from an under-constrained direct fit of $\pmar^{P}$ to the growth over a narrow $\Dc$ range, which would run to the parameter boundary. One dependence we do not fully close: because $\Hr$ enters the saturation fit, the exponent $p$ inherits the token-budget dependence of $\Hr$. The leave-one-out stability above shows $p$ is robust to dropping data points, the joint fit recovers $\Hr$ within $0.002$ bit of the directly measured full-shuffle value, and the interior optimum is sharp; but we have not re-estimated $p$ at a different token budget, and we flag the budget-sensitivity of the precise exponent as a limitation rather than claim it resolved.

\subsection{Provenance of measured, fitted, and derived quantities}

\cref{tab:B1} summarizes the provenance of each quantity: which are design inputs, which are measured, which are fitted, and which relations are derived conditional on the measured fits. We do not claim $0.59$, $p$, or $\Hr$ as first-principles constants.

\begin{table}[H]
\centering
\caption{Provenance of the quantities in the convex-law derivation --- what is measured, fitted, and derived (symbol meanings in \cref{tab:notation}).}
\label{tab:B1}
\begin{tabular}{p{1.8cm}p{6.6cm}p{5.4cm}}
\toprule
symbol & how obtained & nature \\
\midrule
$\rho$ & design input --- the single experimental knob & \textbf{input} \\
$S$ & measured on corrupted tokens (512-token-window adjacency) & \textbf{measured} --- a consequence of $\rho$; experiment-internal, not a universal data property \\
$\Dc$ & plug-in bigram entropy over 2.4M tokens, before training & \textbf{measured} --- the other consequence of $\rho$; the law's data variable \\
$\dlamsq$ & measured post-training (pooled transmission matrices) & \textbf{measured} --- the response \\
$\Hr$ & B-step fit ceiling, cross-checked against the measured full-shuffle $\Dc$ (agree to $0.002$ bit) & \textbf{fitted/measured} anchor \\
$\Delta H$ & B-step fit ($\approx1.82$) & \textbf{fitted} \\
$p$ & B-step fit ($\approx1.69$; interior optimum, leave-one-out stable) & \textbf{empirical fit} --- well-constrained, not first-principles \\
$a_0,a_1$ & training-side linear fit in $S$ ($0.75/0.90$) & \textbf{fitted} \\
$\Czero,\Cone$ & obtained by composition from $a_0,a_1,\Delta H,p$, or fitted directly per $\eta$ ($0.75/0.63$ at the reference $\eta$) & \textbf{response coefficients} --- $\eta$-dependent \\
$0.59$ & $=1/p$ & \textbf{derived} --- the relation is derived; the value is anchored on the empirical $p$ \\
\bottomrule
\end{tabular}
\end{table}

\subsection{\texorpdfstring{Scope of the $\rho$--$S$--$\Dc$ construction}{Scope of the rho-S-D construction}}

The map $\rho\to(S,\Dc)$ is specific to this corruption family. It depends on the stochastic shuffle realization, the clean corpus's statistics, the corruption mechanism, and the estimator budget (the $\Hr$ ceiling and the high-$\rho$ saturation are matched-budget plug-in artifacts; $S$ depends on the 512-token window), so $S$ and $\Dc$ should not be read as universal functions of $\rho$. Along this single positional-shuffle axis the corruption is effectively one-dimensional, so $S$ and $\Dc$ co-vary tightly and $S$ can be eliminated. More general data changes will not reduce to one dimension: redundancy, templating, semantic structure, and higher-order organization can move independently. This is why the present derivation supports the $\Dc$-axis law but also motivates the broader $\Phi(\Dc, R, A, H)$ program of \cref{sec:discussion}.

\clearpage
\section{Data, Models, Code, and Reproducibility}\label{app:repro}

\subsection{Data sources and corruption construction}
The corruption family is built from a WikiText base corpus; the held-out natural corpora are C4 and a deduplicated code corpus, each truncated to a matched $\sim$30M-token budget. Corruption is positional within-sequence shuffling of a fraction $\rho$ of token positions, producing the levels of \cref{sec:method}. The data-side statistics $S$ and $\Dc$ are computed from the resulting token files \emph{before} any training.

\subsection{Model architectures and training setup}
The continued-training runs start from Pythia-70m; the cross-architecture runs train Pythia-70m and a from-scratch Llama-style 70m variant (\cref{sec:robustness}). Optimization --- AdamW with weight decay $\lambda_{\text{wd}}=0.01$, linear warm-up followed by cosine decay, and learning rates spanning $1\text{e-}4$ to $1\text{e-}3$ --- is detailed in \cref{sec:method,sec:robustness}.

\subsection{Weight readout and fitted matrices}
$\lambda$ is the Weibull scale of the pooled transmission-class weight matrices, recorded along each trajectory; $\dlamsq$ subtracts the first-checkpoint value $\lamz^2$ (\cref{sec:method}). The same readout is used at the model and per-layer levels (\cref{sec:robustness}).

\subsection{Scripts for reproducing the law}
The release includes scripts that reproduce the derivation fits (\cref{app:construction}), the exponent-robustness checks, and the self-validation of \cref{app:selfval}, together with the per-run Weibull $\lambda$ trajectories and the pre-computed data-side statistics ($\Dc$ and $S$ for every corruption level and corpus), so the figures reproduce without the raw token files. The corrupted-token files themselves are not redistributed --- they derive from the public WikiText, C4, and CodeParrot corpora --- but the corruption construction script regenerates them from a base corpus for full from-scratch reproduction.

\subsection{Public release}
Code, processed statistics, and fitted trajectories are released at \url{https://github.com/tiexinding/NPM-Weibull-public}, within the unified repository for this line of work.

\clearpage
\section{Supplementary Figures}\label{app:supp}
These figures support the main text and are referenced from it; they are placed here to keep the main narrative to its core figures. They are numbered A1, A2, \ldots, in order of first reference.
\setcounter{figure}{0}
\renewcommand{\thefigure}{A\arabic{figure}}

\begin{figure}[H]\centering
  \includegraphics[width=0.72\linewidth]{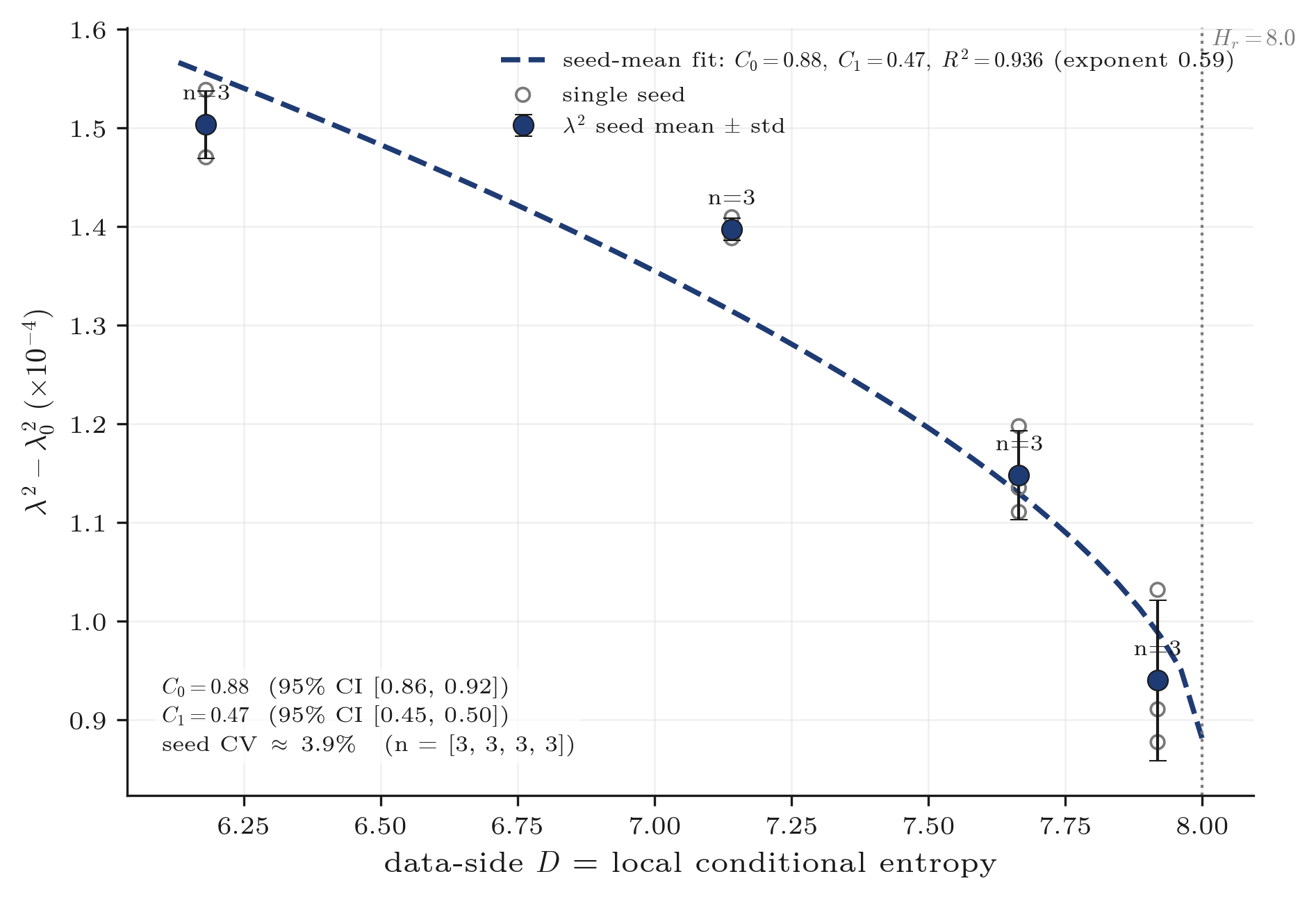}
  \caption{\textbf{Cross-seed replication.} At $\eta=3\times10^{-4}$, $\rho\in\{0,0.2,0.4,0.6\}$ over
    three seeds: growth has coefficient of variation $3.9\%$ (error bars $\pm$std; $n$ per point
    shown), and the seed-mean convex-law fit (exponent $0.59$, as in the main law) gives
    $\Czero=0.88$, $\Cone=0.47$ ($R^2=0.936$) with bootstrap 95\% CIs. The seed variation is small
    relative to the residual scatter.}
  \label{fig:crossseed}
\end{figure}

\begin{figure}[H]\centering
  \includegraphics[width=0.95\linewidth]{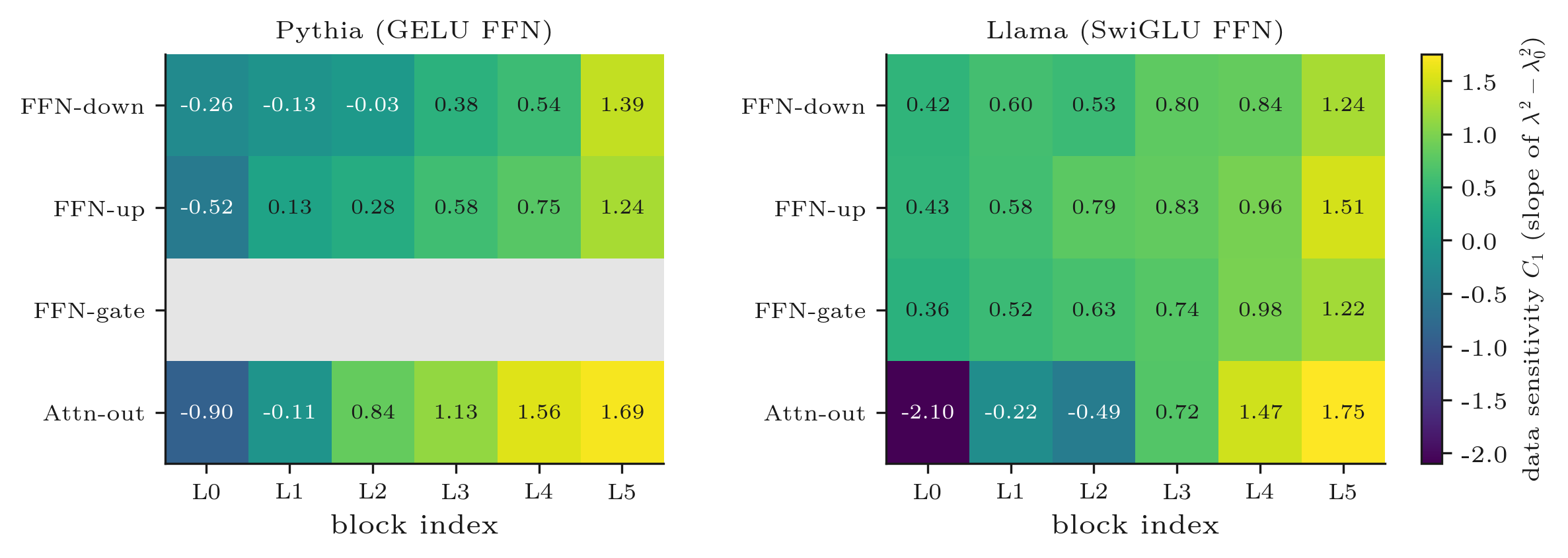}
  \caption{\textbf{Per-block data sensitivity} $\Cone$ (slope of $\dlamsq$), layer $\times$ module,
    for Pythia and Llama. Sequential scale (shared with \cref{fig:arch-lam-maps}); cell values are
    printed, so the few negative entries (shallow blocks) are read directly. Sensitivity rises with
    depth in both; Llama's gated feed-forward responds at every depth, Pythia's only in the deeper
    blocks. Single seed.}
  \label{fig:arch-perlayer}
\end{figure}

\begin{figure}[H]\centering
  \includegraphics[width=0.98\linewidth]{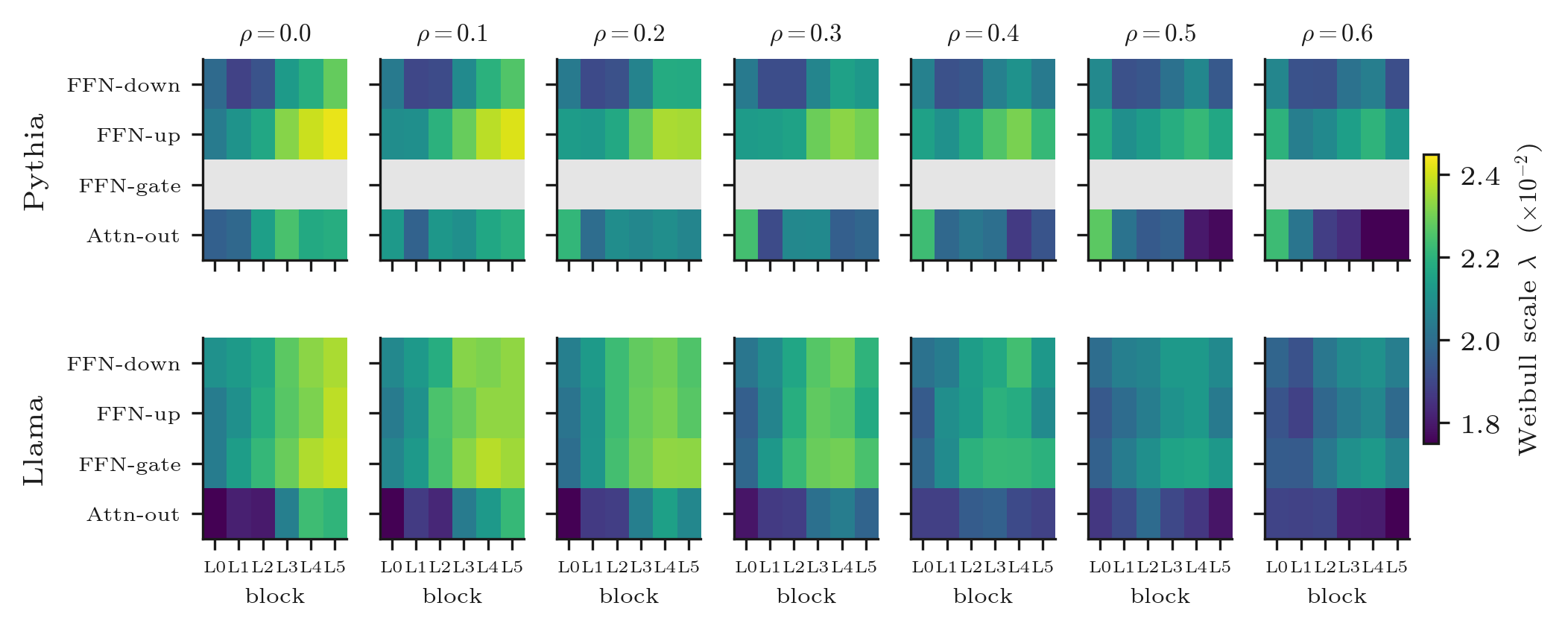}
  \caption{\textbf{Per-$\rho$ Weibull-scale $\lambda$ maps} (layer $\times$ module) for the two
    architectures --- the raw material whose slope along $\rho$ is the $\Cone$ heatmap of
    \cref{fig:arch-perlayer}. Sequential scale; $\lambda$ shrinks as the data is corrupted ($\rho$
    increasing left to right).}
  \label{fig:arch-lam-maps}
\end{figure}

\begin{figure}[H]\centering
  \includegraphics[width=0.78\linewidth]{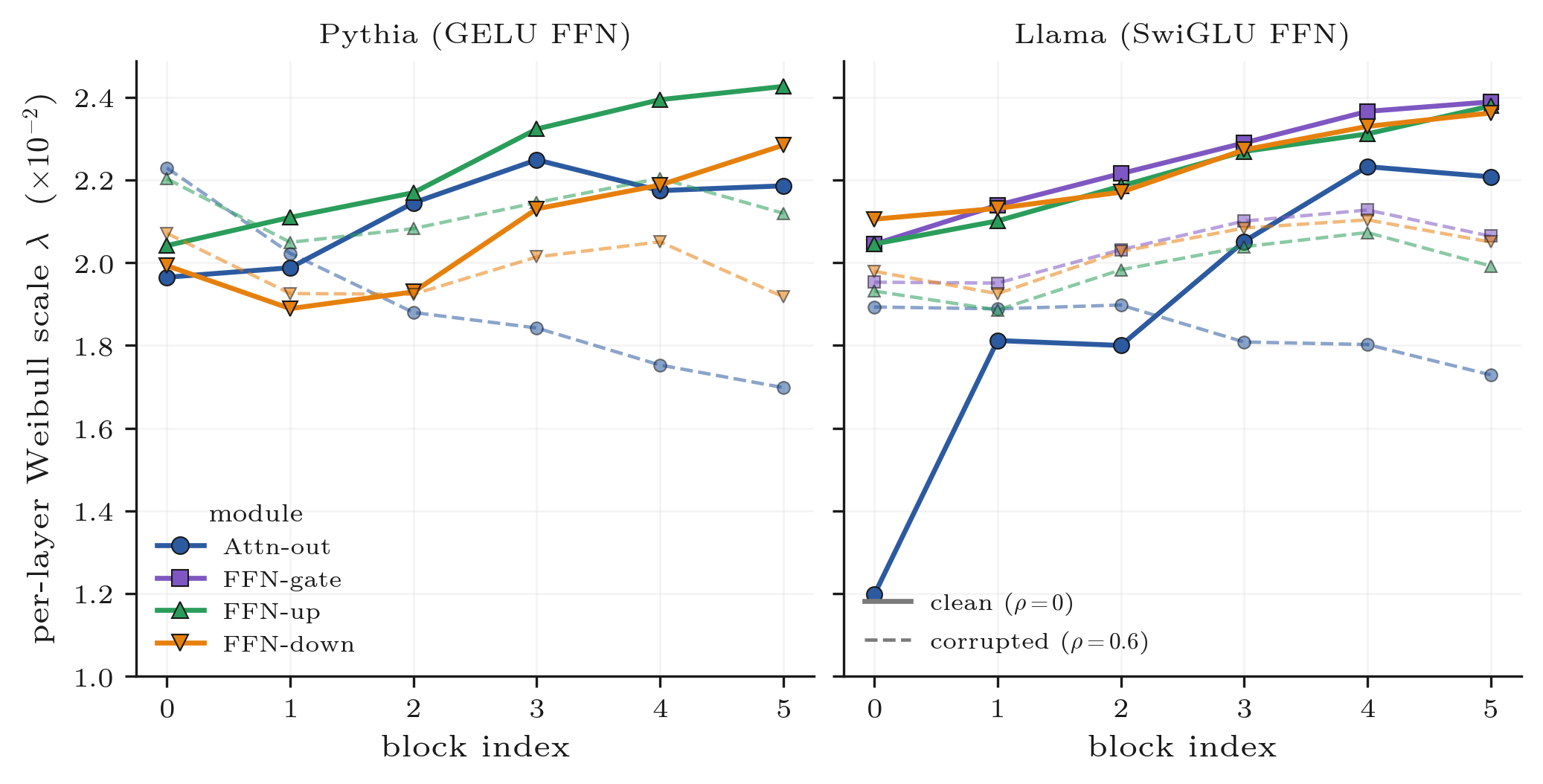}
  \caption{\textbf{End-of-training per-layer $\lambda$ depth profile}, clean ($\rho=0$, solid) vs
    most-corrupted ($\rho=0.6$, dashed), by module, for Pythia and Llama. Single seed.}
  \label{fig:arch-perlayer-lam}
\end{figure}

\begin{figure}[H]\centering
  \includegraphics[width=0.72\linewidth]{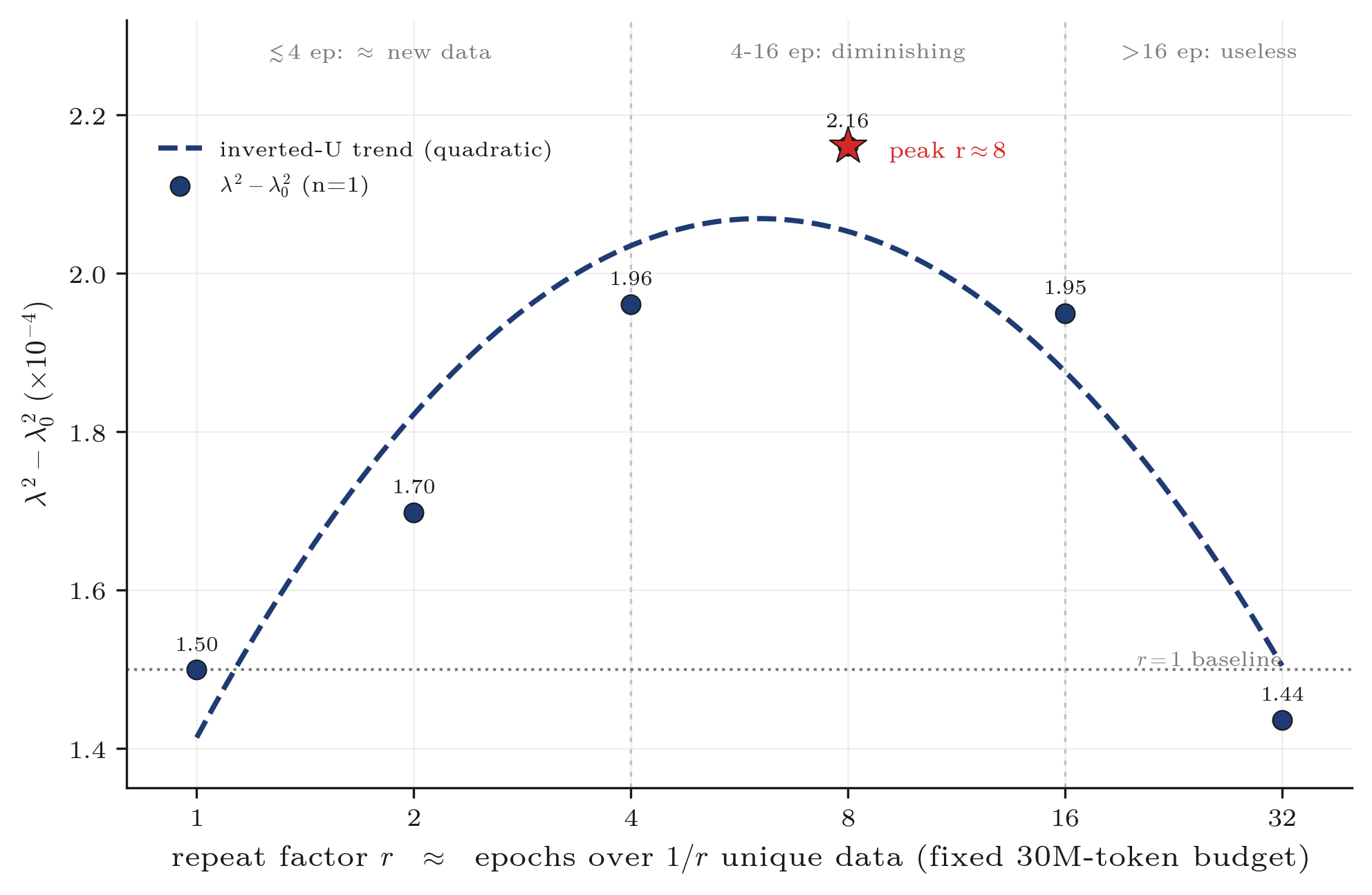}
  \caption{\textbf{Repetition as a second axis (a multi-epoch link; future work).} At fixed token
    budget, a repeat factor $r$ is $\approx r$ epochs over $1/r$ of the data. Growth is non-monotonic
    in $r$ --- rising, peaking near $r\approx8$, then falling --- echoing the diminishing returns of
    repeated data \citep{muennighoff2023scaling}. Single seed; a lead outside the one-dimensional
    law.}
  \label{fig:repeat}
\end{figure}

\end{document}